\pdfoutput=1
\documentclass[11pt]{article}

\newif\ifarxiv
\arxivtrue
\ifarxiv
    \usepackage{acl}
\else
    \usepackage[review]{acl}
\fi
\usepackage{times}
\usepackage{latexsym}
\usepackage{amsmath,amssymb,amsthm}

\usepackage{bbm}
\usepackage{mathabx,mathrsfs}
\usepackage{relsize}
\PassOptionsToPackage{hyphens}{url}\usepackage{hyperref}
\usepackage{xurl}
\usepackage[T1]{fontenc}
\usepackage[utf8]{inputenc}
\usepackage{CJKutf8}
\usepackage{bm}
\usepackage{array}
\usepackage{booktabs}
\usepackage[countmax]{subfloat}
\usepackage{color}
\usepackage{xcolor,colortbl}
\usepackage{mleftright}
\usepackage{caption}
\usepackage{subcaption}
\usepackage{pifont}% http://ctan.org/pkg/pifont
\usepackage{multirow}
\PassOptionsToPackage{hyphens}{url}\usepackage{hyperref}
\usepackage{paralist}
\newcommand\Letter{{\fontfamily{mvs}\fontencoding{U}\selectfont\char66}}
\usepackage{microtype}

\usepackage{inconsolata}

\usepackage{pgfplots}
\usepackage{graphicx}
\usepackage{tcolorbox}
\usepackage{natbib}
\usepackage{tikz}
\usepackage{tikz-dependency}
\usepackage{tikz-qtree}
\usepackage{tikz-layers}
\usepackage{tikzsymbols}
\usepackage{xpinyin}
\usepackage{scalerel}
\usepackage{adjustbox}
\usepackage{tikz-qtree}
\usepackage{nicematrix}
\usepackage{siunitx}
\usepackage{rotating}
\usepgflibrary{arrows.meta}
\usepgfplotslibrary{groupplots}
\usepgfplotslibrary{fillbetween}
\usepgfplotslibrary{statistics}
\usetikzlibrary{fit, calc, plotmarks, decorations.pathreplacing, decorations.markings, positioning, arrows.meta, shapes.geometric, backgrounds, graphs}
\pgfplotsset{compat=1.18}
\usepackage{autobreak}
\usepackage{float}

\usepackage{graphicx}

\pgfdeclarelayer{background}  % declare the background layer
\pgfsetlayers{background,main,above}  % set the layer list

\usepackage{todonotes}
\usepackage{marginnote}

\definecolor{tticblue}{RGB}{0, 94, 184}

\def\hwemoji@insert#1{\scalerel*{\includegraphics[page=#1]{assets/hwemoji-assets.pdf}}{X}}

\newcommand\mailemoji{\hwemoji@insert{3610}}

\definecolor{yanhong}{RGB}{130, 17, 31}
\definecolor{yanlan}{RGB}{20, 74, 116}
\definecolor{json_blue}{RGB}{15, 89, 164}
\definecolor{json_red}{RGB}{192, 44, 53}
\definecolor{figure_green}{RGB}{32, 137, 77}
\definecolor{figure_blue}{RGB}{52, 108, 156}
\definecolor{figure_blue_1}{RGB}{86, 152, 195}
\definecolor{figure_red}{RGB}{192, 44, 53}
\definecolor{figure_orange}{RGB}{250, 126, 35}
\definecolor{table_blue}{RGB}{92, 179, 204}
\definecolor{table_red}{RGB}{238, 63, 77}
\definecolor{table_orange}{RGB}{250, 126, 35}
\definecolor{figure_light_green}{RGB}{198, 223, 200}
\definecolor{figure_light_blue}{RGB}{208, 223, 230}
\definecolor{figure_light_red}{RGB}{192, 44, 53}
\definecolor{figure_light_gray}{RGB}{220, 220, 220}
\definecolor{figure_gray}{RGB}{160, 160, 160}

\newcommand\wz{\phantom{0}}
\newcommand\wm{\phantom{\texttt{-}}}
\newcommand\ewm{\texttt{-}}
\newcommand\ewp{\texttt{+}}
\newcommand\bgood{\hspace{-0.05em}\rlap{\scalebox{0.9}{$^\upharpoonright$}}}
\newcommand\sgood{\hspace{-0.15em}\rlap{\scalebox{0.9}{$^\downharpoonright$}}}

\usepackage{contour}
\usepackage[normalem]{ulem}
\usepackage{soul}

\setuldepth{Berlin}%

\NiceMatrixOptions
       {
         custom-line =
           {
             letter = ; ,
             command = hdashedline ,
             ccommand = cdashedline ,
             tikz = dashed
    } 
}

\pgfdeclarelayer{background}  % declare the background layer
\pgfdeclarelayer{above}       % declare the foreground layer
\pgfsetlayers{background,main,above}  % set the order of the layers (main is the standard layer)

\newcommand \footnotetextonly[1]
{
    \let \backupfootnote \thefootnote
    \let \thefootnote \relax
    \footnotetext{#1}
    \let \thefootnote \backupfootnote
    \let \backupfootnote \imreallyundefinedcommand
}

\title{
Mixture of Small and Large Models for Chinese Spelling Check 
}

\author{Ziheng Qiao,\ \ Houquan Zhou,\ \ Zhenghua Li\rlap{$^{\text{\Letter}}$}\\%
School of Computer Science and Technology, 
Soochow University, China \\%
\texttt{\{zhqiao,hqzhou\}@stu.suda.edu.cn},\ \ 
\texttt{zhli13@suda.edu.cn}}

\begin{document}
\maketitle
% \footnotetextonly{\mailemoji: \textbf{Prof.} Zhenghua Li is the corresponding author.}
\begin{CJK*}{UTF8}{gkai}
    \ifarxiv%
        \footnotetextonly{\!\!\Letter\ Zhenghua Li is the corresponding author.}
    \fi%
    \begin{abstract}

In the era of large language models (LLMs), the Chinese Spelling Check (CSC) task has seen various LLM methods developed, yet their performance remains unsatisfactory. In contrast, fine-tuned BERT-based models, relying on high-quality in-domain data, show excellent performance but suffer from edit pattern overfitting. This paper proposes a novel dynamic mixture approach that effectively combines the probability distributions of small models and LLMs during the beam search decoding phase, achieving a balanced enhancement of precise corrections from small models and the fluency of LLMs. This approach also eliminates the need for fine-tuning LLMs, saving significant time and resources, and facilitating domain adaptation. Comprehensive experiments demonstrate that our mixture approach significantly boosts error correction capabilities, achieving state-of-the-art results across multiple datasets. Our code is available at \url{https://github.com/zhqiao-nlp/MSLLM}.

\end{abstract}

    \section{Introduction}
\label{intro}

The Chinese Spelling Check (CSC) task focuses on identifying and correcting spelling errors in given sentences, as demonstrated in Figure~\ref{fig:examples}. Such errors may lead to comprehension difficulties and adversely affect various natural language processing applications, including machine translation and information retrieval. Given its practical significance, CSC has gained substantial research attention in recent years.

In the era of pre-trained language models, BERT-based approaches \cite{Devlin-etal-2019-bert} have emerged as the dominant solution for the CSC task \cite{zhang-etal-2020-softmasked,Cheng-etal-2020-SpellGCN,Li-etal-2022-SCOPE}. 
Based on the characteristic where the input and output are of equal length for the CSC task, they effectively treat CSC as a character-level classification problem. That is, for each character in the input sentence, they predict whether it needs to be corrected. During the fine-tuning process using in-domain data, either artificial or real-world, these BERT-based models can adeptly capture intricate relationships between edit pairs.
However, this process sometimes leads to overfitting specific edit pairs and generating erroneous sentences.

\begin{figure}[tb!]
    \centering
    \includegraphics[width=0.45\textwidth, trim=0cm 0cm 0cm 0cm, clip]{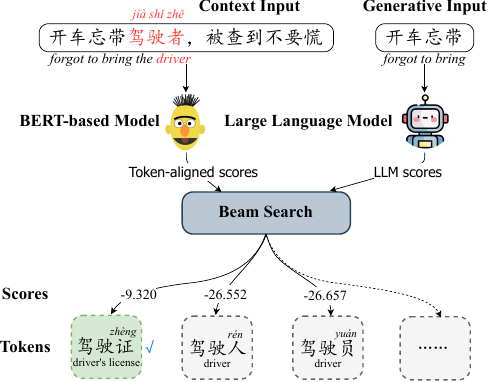} 
    %  left, botton, right,top
    \caption{
        Overview of our approach. The correct sentence is ``开车忘带\textcolor{figure_blue}{驾驶证}，被查到不要慌'' (If you forgot to bring the \textcolor{figure_blue}{driver's license} while driving, don't panic when being checked). 
        }
    \label{fig:big_picture}
\end{figure}

On the other hand, the primary objective of the CSC task is to generate fluent and accurate sentences. 
As a result, generative models are particularly well-suited for this task. With the advent of large language models (LLMs), which boast extensive parameter sizes and vast training datasets, these models exhibit strong cross-domain generalization abilities. 
Researchers have explored various strategies for utilizing LLMs in the CSC task, focusing on whether to fine-tune these models \cite{li-2023-ineffectivenessLLMs,dong-etal-2024-rich}. 
Nevertheless, LLMs often over-polish the text for fluency, choosing expressions they consider better, which leads to inconsistencies between the prediction and the input lengths. 
\citet{li-etal-2024-cllm} attempted to address this issue by employing a character-level tokenization and supervised fine-tuning (SFT) technique. However, this approach requires significant time and resources.

Unlike previous LLM strategies, \citet{zhou-etal-2024-llm-csc} fully utilize the language modeling capabilities of LLMs. They treat open-source LLMs as pure language models and manually design a distortion model to ensure faithfulness between inputs and outputs by leveraging phonetic and glyph similarities. This approach is particularly effective in zero-shot scenarios.

In general, compared to fine-tuned small models, the correction performance of LLM approaches remains unsatisfactory. \citet{liu-etal-2024-arm} tried using an unfine-tuned small model as an arbiter to choose between predictions from both small and large models, yet only achieved minimal improvements. Indeed, due to their different inference mechanisms, BERT-based models and LLMs inherently excel in different aspects of error correction: precision and domain adaptability for BERT-based models, and fluency for LLMs. We believe that a deeper integration of these two models at the inference stage could be a more effective strategy.

Motivated by these insights, this paper presents a novel dynamic mixture approach that strategically integrates a BERT-based model with an LLM. Specifically, we incorporate the probability distribution from the BERT-based model into the LLM's beam search process, thereby preserving the correction capabilities of both models while mitigating the small model's overfitting tendencies through the LLM's robust language modeling. Furthermore, by fine-tuning the small model instead of the LLM, we significantly cut down on the resources and time needed for domain adaptation.

Our contributions are summarized as follows:
\begin{asparaitem}[$\bullet$]
    \item We propose a novel and straightforward approach that combines a BERT-based model with an LLM, leveraging their complementary strengths to further enhance error correction performance.
    \item Our approach does not require fine-tuning the LLM, significantly reducing time and resource costs while preserving its strong generalization capability.
    \item Extensive experiments on multiple mainstream public benchmarks show that our mixture approach substantially boosts correction performance, achieving SOTA results on several datasets.
\end{asparaitem}
    \section{The Basic Approaches}
\label{sec:two_csc_approaches}
Given an input sentence comprising $n$ characters, denoted as $\boldsymbol{x} = x_1 \cdots x_i \cdots x_n$, the objective of a CSC model is to generate a corresponding corrected sentence, represented as $\boldsymbol{y} = y_1 \cdots y_i \cdots y_n$, in which all erroneous characters in $\boldsymbol{x}$ are replaced with the correct ones. In other words, CSC models aim to find an optimal sentence $\boldsymbol{y}$ that maximizes $\texttt{score}(\boldsymbol{x}, \boldsymbol{y})$. Currently, there exist two representative CSC models, i.e., the generative LLM-based models, and the classification-based models.

\subsection{The LLM-based Approach}
\label{sec:approach:generative_model}

\begin{table}[tp!]
    \setlength{\tabcolsep}{12pt}
    \centering
    \scalebox{0.90}{
        \begin{tabular}{lc}
            \toprule
            \textbf{Type}                  & \textbf{Probability}                                                       \\
            \midrule
            \texttt{Identical}             & 0.962 \\
            \midrule
            \texttt{Same \rlap{Pinyin}}    & 0.023 \\
            \texttt{Similar \rlap{Pinyin}} & 0.008 \\
            \texttt{Similar \rlap{Shape}}  & 0.004 \\
            % \texttt{Other Similar} & 家                           & (\textit{\textcolor{figure_blue}{jī}\textcolor{figure_red}{a}})   &wz0.01  \\

            \midrule
            \texttt{Unrelated}             & 0.003 \\
            \bottomrule
        \end{tabular}

    }
    \caption{
        The distribution of the different distortion types extracted from \citet{zhou-etal-2024-llm-csc}.
    }
    \label{tab:types}
\end{table}

Recently, \citet{zhou-etal-2024-llm-csc} proposed a novel prompt-free training-free LLM-based approach for the CSC task. The key is treating LLM as a pure language model. They designed a distortion model (DM) to model the relationships between $\boldsymbol{x}$ and $\boldsymbol{y}$, and more precisely to ensure $\boldsymbol{y}$ is faithful to $\boldsymbol{x}$. 

\begin{equation*}
    \begin{gathered}
        \texttt{score}(\boldsymbol{x}, \boldsymbol{y}) = \log p_{\mathtt{LLM}}(\boldsymbol{y}) + \log p_{\mathtt{DM}}(\boldsymbol{x} \mid \boldsymbol{y}) \\ 
        p_{\mathtt{LLM}}(\boldsymbol{y}) = \prod_{i=1}^{n} p_{\mathtt{LLM}}(y_i\mid \boldsymbol{y}_{<i}) \\
        p_{\mathtt{DM}}(\boldsymbol{x} \mid \boldsymbol{y}) = \prod_{i=1}^{n} p_{\mathtt{DM}}(x_i \mid y_i) \\
        p_{\mathtt{DM}}(x_i \mid y_i) = p(\texttt{type}(x_i,y_i))
    \end{gathered}
\end{equation*}

The LLM component generates a sentence in an auto-regressive manner, and gives us the probability, i.e., $p_{\mathtt{LLM}}(\cdot)$. 

The DM component first classifies each character pair (e.g., $(c_1, c_2)$), into five types, and then obtains the pre-defined corresponding probability, as shown in Table~\ref{tab:types}. 

\subsection{The Classification Approach}
In the pre-trained model era, most mainstream CSC models follow a BERT-based approach \cite{zhang-etal-2020-softmasked,Xu-etal-2021-realise,Li-etal-2022-SCOPE,Liu-etal-2024-ReLM}. These models treat CSC as a local classification problem, i.e., for each character, they determine whether it needs to be corrected and, if so, which character it should be modified to:
\begin{equation*}
    \begin{aligned}
        p_{\mathtt{SM}}(y \mid \boldsymbol{x}, i) = \texttt{softmax}(~ \texttt{MLP}(\boldsymbol{h_i}) ~)[y]
        \label{eq:Distribution}
    \end{aligned}
\end{equation*}
where $\boldsymbol{h_i}$ represents the contextual representation of the $i$-th character obtained from the BERT-based encoder, and $\mathtt{SM}$ is the abbreviation for the small classification model.
By selecting the character with the highest probability at each position, i.e., $y^*  = \mathop{\arg\max}_{y \in \mathcal{V}} p(y \mid \boldsymbol{x}, i)$, where $\mathcal{V}$ represents the vocabulary, we can obtain the final correction result $\boldsymbol{y}$. This classification-based approach demonstrates strong fitting capabilities in specific domains with high-quality training datasets, and it also offers fast decoding speed. 
However, it lacks a global $\texttt{score}(\boldsymbol{x}, \boldsymbol{y})$, which results in the model tending to memorize specific edit pairs, leading to locally optimal solutions.

\section{Our Mixture Approach}
\label{sec:our_approach}

In this work, we propose a straightforward mixture approach to integrate the power of both small and large models. 
On the one hand, the training-free LLM-based approach of \citet{zhou-etal-2024-llm-csc} exhibits remarkable ability in domain generalization. On the other hand, the small classification models can be effectively trained on in-domain labeled data, if available, and thus can dramatically improve in-domain performance. 
\begin{equation*}
\begin{split}
    \texttt{score}(\boldsymbol{x}, \boldsymbol{y}) & = \log p_{\mathtt{LLM}}(\boldsymbol{y}) + \log p_{\mathtt{DM}}(\boldsymbol{x} \mid \boldsymbol{y}) \\
    & + \log p_{\mathtt{SM}}(\boldsymbol{y} \mid \boldsymbol{x}) \\ 
    p_{\mathtt{SM}}(\boldsymbol{y} \mid \boldsymbol{x}) &= \prod_{i=1}^{n} p_{\mathtt{SM}}(y_i \mid \boldsymbol{x}, i)
\end{split}
\end{equation*}

\subsection{Incremental Decomposition} 

In the inference phase, our model follows the LLM component, and produces $\boldsymbol{y}$ from left to right in an auto-regressive manner. 
Thus, we give an incremental decomposition of a partial output sentence as follows: 
% \begin{equation*}
%     \begin{split}
%         \texttt{score}(\boldsymbol{x}, \boldsymbol{y}_{\leq i}) & = \texttt{score}(\boldsymbol{x}, \boldsymbol{y}_{<i})  
%          + \log p_{\mathtt{LLM}}(y_i \mid \boldsymbol{y}_{<i}) \\ & + \log p_{\mathtt{DM}}(x_i \mid y_i) 
%         + \log p_{\mathtt{SM}}(y_i \mid \boldsymbol{x}, i) 
%     \end{split}
% \end{equation*}
\begin{align*}
    \texttt{score}(\boldsymbol{x}, \boldsymbol{y}_{\leq i}) & = \texttt{score}(\boldsymbol{x}, \boldsymbol{y}_{<i}) \\
    & \quad + \log p_{\mathtt{LLM}}(y_i \mid \boldsymbol{y}_{<i}) \\
    & \quad + \log p_{\mathtt{DM}}(x_i \mid y_i) \\
    & \quad + \log p_{\mathtt{SM}}(y_i \mid \boldsymbol{x}, i)
\end{align*}

\subsection{Token-based Generation and Beam Search Decoding}

Current LLMs usually generate a sentence token by token, i.e., using tokens as the basic units. Therefore, the output sentence can also be denoted as $\boldsymbol{y} = \boldsymbol{t}_1 \cdots \boldsymbol{t}_{k} \cdots \boldsymbol{t}_{o}$, where a token is composed of $\ell \geq 1$ characters, i.e., $\boldsymbol{t}_k = y_{i-\ell+1} \dots y_i$.

Moreover, we follow \citet{zhou-etal-2024-llm-csc} and employ their proposed faithfulness reward to further encourage that $\boldsymbol{y}$ retains the same meaning as $\boldsymbol{x}$. 

\begin{table*}[tb!]
    \setlength{\tabcolsep}{3pt}
    \renewcommand{\arraystretch}{1}
    \centering
    \scalebox{1.0}{
        \begin{NiceTabular}{lcc>{\columncolor{figure_light_red!6}}c|cc>{\columncolor{figure_light_red!6}}c|cc>{\columncolor{figure_light_red!6}}c|cc>{\columncolor{figure_light_red!6}}c|cc>{\columncolor{figure_light_red!6}}c}
            \toprule
            \rowcolor[gray]{1.0}
            \Block[l]{2-1}{\textbf{Model}} & \Block[c]{1-3}{\textbf{rSIGHANs}} & & & \Block[c]{1-3}{\textbf{CSCD-NS}} & & & \Block[c]{1-3}{\textbf{MCSCSet}} & & & \Block[c]{1-3}{\textbf{ECSpell}} & & & \Block[c]{1-3}{\textbf{LEMON}} & & \\
            \rowcolor[gray]{1.0} & S-F\bgood & C-F\bgood & FPR\sgood & S-F\bgood & C-F\bgood & FPR\sgood & S-F\bgood & C-F\bgood & FPR\sgood & S-F\bgood & C-F\bgood & FPR\sgood & S-F\bgood & C-F\bgood & FPR\sgood \\

            \midrule
            \rowcolor[gray]{.95}
            \Block[c]{1-16}{\texttt{Previous SOTAs}} & & & & & & & & & & & & & & & \\
            \Block[l]{1-1}{\texttt{BERT}\rlap{$^\dagger$}} & 70.8 & \textbf{81.9} & 12.1 & \textbf{77.4} & 79.3 & 10.9 & \textbf{87.6} & \textbf{94.0} & 1.8 & 91.0 & 94.6 & \textbf{3.8} & 48.1 & 49.3 & 13.1 \\
            \Block[l]{1-1}{\texttt{ReLM}\rlap{$^\dagger$}} & \textbf{71.8} & 81.8 & \textbf{11.5} & 77.3 & \textbf{79.7} & \textbf{10.5} & 87.0 & 93.7 & \textbf{1.6} & \textbf{92.3} & \textbf{94.9} & 7.4 & \textbf{50.6} & \textbf{51.6} & \textbf{11.7} \\

            \midrule
            \rowcolor[gray]{.95}
            \Block[c]{1-16}{\texttt{LLMs} \cite{zhou-etal-2024-llm-csc}} & & & & & & & & & & & & & & & \\
            \Block[l]{1-1}{\texttt{Baichuan2}} & \textbf{59.1} & \textbf{70.1} & \textbf{15.4} & \textbf{62.7} & \textbf{65.7} & \textbf{16.8} & \textbf{67.2} & \textbf{78.0} & \textbf{2.0} & \textbf{85.4} & \textbf{90.2} & \textbf{5.1} & \textbf{53.2} & \textbf{56.7} & \textbf{9.9} \\
            \Block[l]{1-1}{\texttt{Qwen2.5}} & 55.6 & 68.6 & 17.9 & 58.6 & 62.6 & 23.5 & 63.3 & 74.1 & 3.2 & 81.3 & 88.1 & 6.6 & 48.6 & 53.5 & 12.8 \\

            \Block[l]{1-1}{\texttt{IL2.5}} & 52.5 & 66.2 & 20.1 & 55.3 & 59.1 & 26.5 & 51.9 & 62.8 & 5.3 & 80.3 & 87.0 & 6.9 & 45.3 & 50.0 & 15.6 \\
            \midrule

            \rowcolor[gray]{.95}
            \Block[c]{1-16}{\texttt{Ours}} & & & & & & & & & & & & & & & \\
            % \Block[l]{1-1}{\texttt{ARM}} & -- & -- & -- & -- & -- & -- & -- & -- & -- & -- & -- & -- & 32.2 & -- & -- \\
            % \hdashedline
            \Block[l]{1-1}{\texttt{BERT\;+\;BC2}} & 72.5 & 83.4 & 7.4 & 78.1 & 79.1 & 10.1 & 91.2 & 96.0 & 1.4 & 94.4 & 96.3 & \textbf{1.7} & 56.4 & 58.2 & 6.9 \\
            \Block[l]{1-1}{\texttt{BERT\;+\;QW2.5}} & 73.6 & \textbf{83.9} & \textbf{7.0} & 76.8 & 78.6 & 11.3 & 91.6 & 95.9 & 1.6 & 94.0 & 95.5 & 2.7 & 56.1 & 58.7 & 10.1 \\
            \Block[l]{1-1}{\texttt{BERT\;+\;IL2.5}} & 72.8 & 83.8 & 7.9 & 77.3 & 78.3 & 11.5 & 90.7 & 95.7 & 1.7 & 94.3 & 94.9 & 2.2 & 53.8 & 55.1 & 6.7 \\
            \hdashedline

            \Block[l]{1-1}{\texttt{ReLM\;+\;BC2}} & 73.9 & 83.0 & 9.2 & \textbf{79.9} & \textbf{81.8} & \textbf{8.4} & 91.7 & 96.2 & \textbf{1.1} & \textbf{97.1} & \textbf{98.3} & 2.1 & \textbf{61.0} & \textbf{61.6} & \textbf{5.6} \\
            \Block[l]{1-1}{\texttt{ReLM\;+\;QW2.5}} & 73.9 & 82.6 & 9.1 & 78.9 & 79.8 & 10.0 & \textbf{92.1} & \textbf{96.3} & 1.3 & 96.6 & 97.8 & 3.0 & 60.5 & 61.4 & 7.9 \\

            \Block[l]{1-1}{\texttt{ReLM\;+\;IL2.5}} & \textbf{74.4} & 83.6 & 9.5 & 79.3 & 81.0 & 10.1 & 90.4 & 95.4 & 1.5 & 96.7 & 98.0 & 2.6 & 58.9 & 60.2 & 7.9 \\
            \bottomrule

        \end{NiceTabular}
    }
    \caption{
        Sentence- and character-level results on the mainstream CSC test sets. $F_1$ and FPR scores are reported (\%). 
        \texttt{BC2} stands for \texttt{Baichuan2}, \texttt{QW2.5} for \texttt{Qwen2.5}, and \texttt{IL2.5} for \texttt{InternLM2.5}.
        All LLMs use the 7B model size version.
        Note that the performance metrics for rSIGHANs, ECSpell, and LEMON are presented as macro averages.
        ``$\dagger$'' indicates that the models are pre-trained on 34 million synthetic data and fine-tuned with MFT strategy (LEMON cannot be further fine-tuned due to the lack of in-domain training data). 
    }
    \label{tab:main_results}
\end{table*}

\paragraph{The full model.} Combining the above two factors, we give the incremental decomposition of our full model. 

\begin{equation}
    \begin{aligned}
        & \texttt{score}(\boldsymbol{x}, \boldsymbol{y}_{\le i}=\boldsymbol{t}_{\le k} ) = \texttt{score}(\boldsymbol{x}, \boldsymbol{t}_{< k} ) \\
        & \ + \log p_{\mathtt{LLM}}(\boldsymbol{t_k} \mid \boldsymbol{t}_{< k}) \\
        & \ + (1+H_{\mathtt{LLM}}(\cdot))\times\left(\begin{array}{c}
            \alpha \times \log p_{\mathtt{DM}}(\boldsymbol{x}, i\mid \boldsymbol{t}_k) \\
            + \\
            \beta \times  \log p_{\mathtt{SM}}(\boldsymbol{t}_k\mid \boldsymbol{x}, i) \end{array}\right)
    \\ 
    \end{aligned}
    \label{eq:final_score}
\end{equation}
where $\alpha$ and $\beta$ are the weights of the distortion component in the generative LLM-based model and the small BERT-based model, respectively; $H_{\mathtt{LLM}}(\cdot)$ corresponds to the faithfulness reward of \citet{zhou-etal-2024-llm-csc}, and represents the entropy of $\smash{p_{\mathtt{LLM}}(\boldsymbol{t}_k \mid \boldsymbol{t}_{< k})}$, i.e., the probability distribution of the LLM component regarding the generation of $\boldsymbol{t}_k$. Higher entropy means the LLM is more uncertain about the selection of the next token, and thus the other two components obtain higher weights. 

The token-level formulations for the DM component and the small model are defined as follows:
\begin{equation*}
    \begin{gathered}
        p_{\mathtt{DM}}(\boldsymbol{x}, i\mid \boldsymbol{t}_k) = \prod_{j=1}^{\ell}{p_{\mathtt{DM}}( x_{i-\ell+j} \mid \boldsymbol{t}_{k}[j-1])} \\
        p_{\mathtt{SM}}(\boldsymbol{t}_k\mid \boldsymbol{x}, i) = \prod_{j=1}^{\ell}{p_{\mathtt{SM}}( \boldsymbol{t}_{k}[j-1] \mid \boldsymbol{x}, i-\ell+j)} 
    \end{gathered}
\end{equation*}

For multi-word tokens, the value of $p_{\mathtt{DM}}(\cdot)$ is computed as the product of the probabilities of all characters within the token. In the method of \citet{zhou-etal-2024-llm-csc}, the weight $\alpha$ of the distortion component is set to 1. The computation of $p_{\mathtt{SM}}(\cdot)$ involves segmenting a LLM token into multiple small-model tokens and using the probability distribution from the small model to derive the score of the corresponding multi-word token, thereby achieving token-level alignment.

\paragraph{Beam search.} 
During inference, we follow \citet{zhou-etal-2024-llm-csc} and employ beam search, in order to explore a larger search space. At each decoding step, we retain the top $K$ candidates with the highest scores, and based on them build candidates for the next step.
    \section{Experimental Setup}
\subsection{Datasets}
\paragraph{Chinese Learner Texts.}
Following the conventions of previous works, we employ the \textbf{SIGHANs} datasets \cite{wu-etal-2013-chinese,yu-etal-2014-overview,tseng-etal-2015-introduction} as our benchmarks, which are derived from Chinese learner texts. We utilize a revised version of SIGHANs \cite{Yang-etal-2023-rSIGHAN}, which was manually annotated for errors and noise present in the original SIGHANs. We refer to this version as \textbf{rSIGHANs}. 
In the training stage, we use Wang271K~\cite{wang-etal-2018-hybrid} + SIGHANs as our training set.

\paragraph{Chinese Native-speaker Texts.}
Due to the lack of domain diversity in the SIGHAN datasets, we further conduct experiments on more diverse datasets, including \textbf{LEMON} \cite{Wu-etal-2023c-LEMON}, \textbf{ECSpell} \cite{Lv-etal-2023-ECSpell}, \textbf{CSCD-NS} \cite{hu-etal-2024-cscd}, and \textbf{MCSCSet} \cite{jiang-etal-2022-mcscset}, all of which were written by native speakers. Notably, LEMON includes test sets from seven different domains but does not provide in-domain training sets, making it an ideal benchmark for evaluating a model's cross-domain generalization capability. Detailed information about all datasets can be found in Appendix~\ref{sec:appendix-datasets}.

\subsection{Baseline Models}
We select several BERT-based models and LLMs as our baselines.
\paragraph{Small BERT-based Models.}
For benchmarks across various domains, we select BERT \cite{Devlin-etal-2019-bert} and ReLM \cite{Liu-etal-2024-ReLM} for experiments. Additionally, on the rSIGHAN15 test set, we select several mainstream SOTA models for experiments and report the results, such as ReaLiSe \cite{Xu-etal-2021-realise} and SCOPE \cite{Li-etal-2022-SCOPE}.
Detailed information about all baselines can be found in Appendix~\ref{sec:appendix-baselines}.

\paragraph{Open-source LLMs.}
We use \texttt{Baichuan2} \cite{yang-etal-2023-baichuan2}, \texttt{Qwen2.5} \cite{bai-etal-2023-qwen}, and \texttt{InternLM2.5} \cite{cai-etal-2024-internlm2} as our open-source LLMs for experiments. All LLMs use the \underline{base} version.
In the main experiments, we fix the model size to 7B. For comprehensive and robust experiments, in Section~\ref{sec:ablation:family_size}, we select LLMs with different sizes ranging from 0.5B to 20B.

\begin{figure}[tb!]
    \captionsetup[subfigure]{skip=2pt}
    \begin{center}
        
        \subfloat[Correct errors from ReLM and LLM]{
            \centering
            \scalebox{0.96}{
                \begin{tikzpicture}[
                        font=\scriptsize,
                        anchor point/.style={
                                draw=none,
                                circle,
                                inner sep=1.5pt,
                            },
                        word/.style={
                                font=\footnotesize,
                                anchor=base,
                            },
                        bg/.style={
                                fill opacity=0.6,
                                rounded corners=3pt,
                                inner sep=1.2pt,
                                minimum width=0.98\linewidth,
                                minimum height=7 0.0,
                            },
                    ]

                    % anchor for four corners
                    \node[anchor point] (a west) at (-0.48\linewidth, -2.5) {};
                    \node[anchor point] (a east) at (0.48\linewidth, -2.5) {};
                    \node[anchor point] (a center) at (0, -2.5) {};
                    % source sentence
                    \node[word, inner sep=0pt, anchor=north west] (source title) at (a west) {\textbf{\texttt{Input:}}};
                    \node[word, anchor=base west] (source) at ($(source title.base east) + (0.6, 0)$) {\textcolor{figure_red}{水饺}和新鲜的空气都很重要。};
                    \node[word, anchor=north west, font=\scriptsize, align=left] (source gross) at ($(source.south west) + (0.0, 0.2)$) {\quad \textit{\textcolor{figure_red}{Dumplings} and fresh air are both important.}};
                    % reference sentence
                    \node[word, inner sep=0pt, anchor=north west] (reference title) at ($(source title.south west) + (0, -0.45)$) {\textbf{\texttt{Reference:}}};
                    \node[word, anchor=base west] (reference) at ($(reference title.base-|source.west) + (0, -0.0)$) {\textcolor{figure_red}{水饺} \ding{212} \textcolor{figure_blue}{睡觉} \scriptsize{(\textit{\textcolor{figure_red}{shuǐjiǎo}} \ding{212} \textit{\textcolor{figure_blue}{shuìjiào, sleep}})}};
                    % predict sentence
                    \node[word, inner sep=0pt, anchor=north west] (predict title) at ($(reference title.south west) + (0, -0.2)$) {\textbf{\texttt{ReLM:}}};
                    \node[word, anchor=base west] (predict) at ($(predict title.base-|source.west) + (0, -0.0)$) {\textcolor{figure_red}{水饺} \ding{212} \textcolor{figure_red}{水}\textcolor{figure_blue}{觉} \scriptsize{(\textit{\textcolor{figure_red}{shuǐjiǎo}} \ding{212} \textit{\textcolor{figure_red}{shuǐ}\textcolor{figure_blue}{jiào}\textcolor{figure_red}{, water sleep}})}};
                    % predict sentence
                    \node[word, inner sep=0pt, anchor=north west] (predict title) at ($(predict title.south west) + (0, -0.2)$) {\textbf{\texttt{LLM:}}};
                    \node[word, anchor=base west] (predict) at ($(predict title.base-|source.west) + (0, -0)$) {\texttt{NONE}};
                    % \node[word, anchor=base west] (predict) at ($(predict title.base-|source.west) + (0, -0.0)$) {\textcolor{figure_red}{水饺} \ding{212} \textcolor{figure_red}{水觉} \scriptsize{(\textit{\textcolor{figure_red}{shuǐjiǎo}} \ding{212} \textit{\textcolor{figure_red}{shuǐjiào, water sleep}})}};
                    % predict sentence
                    \node[word, inner sep=0pt, anchor=north west] (predict title gpt4) at ($(predict title.south west) + (0, -0.2)$) {\textbf{\texttt{ReLM+LLM:}}};
                    \node[word, anchor=base west] (predict gpt4) at ($(predict title gpt4.base-|source.west) + (0, -0.0)$) {\textcolor{figure_red}{水饺} \ding{212} \textcolor{figure_blue}{睡觉} \scriptsize{(\textit{\textcolor{figure_red}{shuǐjiǎo}} \ding{212} \textit{\textcolor{figure_blue}{shuìjiào, sleep}})}};
                    \begin{scope}[on background layer]
                        \node[fill=lightgray!30, bg, anchor=north] at ($(a center) + (0.0, 0.15)$) {};
                    \end{scope}
                \end{tikzpicture}
            }
            \label{fig:examples:correct_both}
        }\\[3pt]%
        \subfloat[Correct errors from LLM]{
            \centering
            \scalebox{0.96}{
                \begin{tikzpicture}[
                        font=\scriptsize,
                        anchor point/.style={
                                draw=none,
                                circle,
                                inner sep=1.5pt,
                            },
                        word/.style={
                                font=\footnotesize,
                                anchor=base,
                            },
                        bg/.style={
                                fill opacity=0.6,
                                rounded corners=3pt,
                                inner sep=1.2pt,
                                minimum width=0.98\linewidth,
                                minimum height=76.0,
                            },
                    ]
                    % anchor for four corners
                    \node[anchor point] (a west) at (-0.48\linewidth, 0) {};
                    \node[anchor point] (a east) at (0.48\linewidth, 0) {};
                    \node[anchor point] (a center) at (0, 0) {};
                    \node[word, inner sep=0pt, anchor=north west] (source title) at (a west) {\textbf{\texttt{Input:}}};
                    \node[word, anchor=base west] (source) at ($(source title.base east) + (0.6, 0)$) {开车忘带驾驶\textcolor{figure_red}{者},被查到不要慌，...};
                    \node[word, anchor=north west, font=\scriptsize, align=left] (source gross) at ($(source.south west) + (0.0, 0.2)$) {\quad \textit{If you forgot to bring the \textcolor{figure_red}{driver} while driving, }\\
                    [-2pt]\textit{don't panic when being checked,...}};
                    % reference sentence
                    \node[word, inner sep=0pt, anchor=north west] (reference title) at ($(source title.south west) + (0, -0.65)$) {\textbf{\texttt{Reference:}}};
                    \node[word, anchor=base west] (reference) at ($(reference title.base-|source.west) + (0, -0)$) {驾驶\textcolor{figure_red}{者} \ding{212} 驾驶\textcolor{figure_blue}{证}\scriptsize{(\textit{\textcolor{figure_red}{zhě}} \ding{212} \textit{\textcolor{figure_blue}{zhèng, driver's license}})}};
                    % predict sentence
                    \node[word, inner sep=0pt, anchor=north west] (predict title) at ($(reference title.south west) + (0, -0.2)$) {\textbf{\texttt{ReLM:}}};
                    \node[word, anchor=base west] (predict) at ($(predict title.base-|source.west) + (0, -0)$) {驾驶\textcolor{figure_red}{者} \ding{212} 驾驶\textcolor{figure_blue}{证}\scriptsize{(\textit{\textcolor{figure_red}{zhě}} \ding{212} \textit{\textcolor{figure_blue}{zhèng, driver's license}})}};
                    % predict sentence
                    \node[word, inner sep=0pt, anchor=north west] (predict title) at ($(predict title.south west) + (0, -0.2)$) {\textbf{\texttt{LLM:}}};
                    \node[word, anchor=base west] (predict gpt4) at ($(predict title.base-|source.west) + (0, 0)$) {\texttt{NONE}};
                    % predict sentence
                    \node[word, inner sep=0pt, anchor=north west] (predict title relm_llm) at ($(predict title.south west) + (0, -0.2)$) {\textbf{\texttt{ReLM+LLM:}}};
                    \node[word, anchor=base west] (predict gpt4) at ($(predict title relm_llm.base-|source.west) + (0, 0)$) {驾驶\textcolor{figure_red}{者} \ding{212} 驾驶\textcolor{figure_blue}{证}\scriptsize{(\textit{\textcolor{figure_red}{zhě}} \ding{212} \textit{\textcolor{figure_blue}{zhèng, driver's license}})}};;

                    \begin{scope}[on background layer]
                        \node[fill=lightgray!30, bg, anchor=north] at ($(a center) + (0.0, 0.15)$) {};
                    \end{scope}
                \end{tikzpicture}
            }
            \label{fig:examples:correct_llm}
        }\\[3pt]%
        \subfloat[Correct errors from ReLM]{
            \centering
            \scalebox{0.96}{
                \begin{tikzpicture}[
                        font=\scriptsize,
                        anchor point/.style={
                                draw=none,
                                circle,
                                inner sep=1.5pt,
                            },
                        word/.style={
                                font=\footnotesize,
                                anchor=base,
                            },
                        bg/.style={
                                fill opacity=0.6,
                                rounded corners=3pt,
                                inner sep=1.2pt,
                                minimum width=0.98\linewidth,
                                minimum height=76.0,
                            },
                    ]
                    \node[anchor point] (a west) at (-0.48\linewidth, 0) {};
                    \node[anchor point] (a east) at (0.48\linewidth, 0) {};
                    \node[anchor point] (a center) at (0, 0) {};
                    \node[word, inner sep=0pt, anchor=north west] (source title) at (a west) {\textbf{\texttt{Input:}}};
                    \node[word, anchor=base west] (source) at ($(source title.base east) + (0.6, 0)$) {不仅赖账，还提无\textcolor{figure_red}{力}要求！};
                    \node[word, anchor=north west, font=\scriptsize, align=left] (source gross) at ($(source.south west) + (0.0, 0.2)$) {\quad \textit{Not only do they refuse to pay, but they also }\\
                    [-2pt]\textit{make \textcolor{figure_red}{powerless} demands!}};
                    \node[word, inner sep=0pt, anchor=north west] (reference title) at ($(source title.south west) + (0, -0.65)$) {\textbf{\texttt{Reference:}}};
                    \node[word, anchor=base west] (predict gpt4) at ($(reference title.base-|source.west) + (0, 0)$) {无\textcolor{figure_red}{力} \ding{212} 无\textcolor{figure_blue}{理}\scriptsize{(\textit{\textcolor{figure_red}{wúlì}} \ding{212} \textit{\textcolor{figure_blue}{wúlǐ, unreasonable}})}};
                    \node[word, inner sep=0pt, anchor=north west] (predict title) at ($(reference title.south west) + (0, -0.2)$) {\textbf{\texttt{ReLM:}}};
                    \node[word, anchor=base west] (predict) at ($(predict title.base-|source.west) + (0, -0)$) {\textcolor{figure_blue}{提} \ding{212} \textcolor{figure_red}{是}\scriptsize{(\textit{\textcolor{figure_blue}{tí}} \ding{212} \textit{\textcolor{figure_red}{shì, are}})}};
                    \node[word, inner sep=0pt, anchor=north west] (predict title) at ($(predict title.south west) + (0, -0.2)$) {\textbf{\texttt{LLM:}}};
                    \node[word, anchor=base west] (predict gpt4) at ($(predict title.base-|source.west) + (0, -0)$) {无\textcolor{figure_red}{力} \ding{212} 无\textcolor{figure_blue}{理}\scriptsize{(\textit{\textcolor{figure_red}{wúlì}} \ding{212} \textit{\textcolor{figure_blue}{wúlǐ, unreasonable}})}};
                    \node[word, inner sep=0pt, anchor=north west] (predict title relm_llm) at ($(predict title.south west) + (0, -0.2)$) {\textbf{\texttt{ReLM+LLM:}}};
                    \node[word, anchor=base west] (predict gpt4) at ($(predict title relm_llm.base-|source.west) + (0, -0)$) {无\textcolor{figure_red}{力} \ding{212} 无\textcolor{figure_blue}{理}\scriptsize{(\textit{\textcolor{figure_red}{wúlì}} \ding{212} \textit{\textcolor{figure_blue}{wúlǐ, unreasonable}})}};;
                    \begin{scope}[on background layer]
                        \node[fill=lightgray!30, bg, anchor=north] at ($(a center) + (0.0, 0.15)$) {};
                    \end{scope}
                \end{tikzpicture}
            }
            \label{fig:examples:correct_relm}
        }
    \end{center}
    \caption{Cases from rSIGHAN15 and LEMON-\textit{New} test sets. The LLM used is \texttt{Baichuan2(7B)}.}
    \label{fig:examples}
\end{figure}
\paragraph{Mixture Strategy.}
ARM \cite{liu-etal-2024-arm} attempts to make trade-offs between the correction results of small models and \texttt{GPT-3.5-Turbo}.\footnote{\url{https://platform.openai.com}} Due to the lack of open-source code and differences in experimental settings on LEMON, we only list their results in Table~\ref{tab:sighan15} and Table~\ref{tab:subsets_sf}.

\subsection{Evaluation Metrics}
Following the mainstream evaluation metrics for CSC tasks, we report the Precision (P), Recall (R), and $F_1$ scores of the correction subtask at both the sentence- and character-level, denoted as S-P/R/F and C-P/R/F, respectively. To comprehensively evaluate the model's correction capability, we also include the false positive rate (FPR) as an additional metric.

\subsection{Implementation Details}
For ReaLiSe, we employ the pre-trained checkpoint available from its official GitHub repository.\footnote{\url{https://github.com/DaDaMrX/ReaLiSe}} For SCOPE, we adopt their official implementation for fine-tuning.\footnote{\url{https://github.com/jiahaozhenbang/SCOPE}} Both BERT and ReLM utilize the framework from \citet{Liu-etal-2024-ReLM}, with their pre-trained models (trained on a corpus of 34 million synthetic sentences) being fine-tuned in our experiments.\footnote{\url{https://github.com/gingasan/lemon}} During training, we set the batch size to 128 and the learning rate to $3\times10^{-5}$, while incorporating the MFT strategy \cite{Wu-etal-2023c-LEMON}. All experiments are conducted on an NVIDIA A100-PCIE-40GB GPU.

\section{Main Results}
Table~\ref{tab:main_results} presents experimental results across five benchmark datasets: rSIGHANs, CSCD-NS, MCSCSet, ECSpell, and LEMON. When integrated with LLMs, nearly all models exhibit substantial enhancements across sentence- and character-level correction metrics (Precision, Recall, and $F_1$). We also provide the results of the \textbf{original SIGHAN15} dataset in Appendix~\ref{sec:appendix-sighan15}.

To evaluate cross-domain generalization capability, we specifically test on ECSpell and LEMON datasets. Our approach demonstrates consistent performance gains on both test sets, with detailed sub-domain results provided in Appendix~\ref{sec:appendix-detailed-results}. 
Notably, ReLM integrated with \texttt{Baichuan2} achieves remarkable improvements: a 10.4\% average increase in sentence-level $F_1$ across all seven LEMON domains, and 4.8\% enhancement on ECSpell.
During our experiments, we discovered that ECSpell has a serious issue of target sentence leakage, leading to overly high performance of fine-tuned small models.
We will explain this issue and present the results on the cleaned ECSpell dataset in Appendix~\ref{sec:appendix-cleaned-ecspell}.

Due to different experimental setups, we also list another mixture strategy, ARM \cite{liu-etal-2024-arm}, in Appendix~\ref{sec:appendix-more-experiments}. In comparison, our method achieves greater improvements across all datasets.

\begin{table}[tb!]
    \setlength{\tabcolsep}{4.0pt}
    \renewcommand{\arraystretch}{0.9}
    \centering
    \scalebox{0.85}{
        \begin{NiceTabular}{lrp{2.2em}p{2.2em}|p{2.2em}p{2.2em}|p{2.2em}p{2.2em}}
            \toprule
            \rowcolor[gray]{1.0}
            \Block[l]{2-1}{\textbf{LLM}} & \Block[l]{2-1}{\textbf{Size}}                          & \Block[c]{1-2}{\textbf{rSIGHAN15}} &               & \Block[c]{1-2}{\textbf{ECSpell}-\textit{Odw}} &               & \Block[c]{1-2}{\textbf{LEMON}-\textit{Nov}} &       \\
            \rowcolor[gray]{1.0}
                                            &                           & S-F\bgood     & C-F\bgood       & S-F\bgood     & C-F\bgood     & S-F\bgood     & C-F\bgood    \\
            \midrule
            % \Block[l]{2-1}{\texttt{\shortstack[l]{ReLM\\ \ +\ BC2}}}
            \Block[l]{2-1}{\texttt{BC2}}
                                            & \texttt{7B}  & \textbf{79.6} & \textbf{85.4} & 96.5 & 98.2 & 50.0 & 50.4 \\
                                            & \texttt{13B} & 77.9 & 84.6 & \textbf{96.7} & \textbf{98.3} & \textbf{50.8} & \textbf{50.9} \\
            \midrule
            % \Block[l]{5-1}{\texttt{\shortstack[l]{ReLM\\ \ +\ QW2.5}}}
            \Block[l]{5-1}{\texttt{QW2.5}}
                                            & \texttt{0\!.\!5B}  & 78.4 & 84.2 & 95.7 & 97.3 & 45.4 & 46.3 \\
                                            & \texttt{1\!.\!5B}  & 78.2 & 83.3 & 95.7 & 97.6 & 48.1 & 49.0 \\
                                            & \texttt{3B}        & \textbf{79.3} & 83.7 & 96.7 & 98.1 & 48.8 & 49.9 \\
                                            & \texttt{7B}        & 77.9 & 84.2 & 95.9 & 97.7 & 49.4 & 50.6 \\
                                            & \texttt{14B}       & 78.6 & \textbf{84.3} & \textbf{97.3} & \textbf{98.5} & \textbf{50.5} & \textbf{51.5} \\
            \midrule
            % \Block[l]{3-1}{\texttt{\shortstack[l]{ReLM\\ \ +\ IL2.5}}}
            \Block[l]{3-1}{\texttt{IL2.5}}
                                            & \texttt{1\!.\!8B} & 79.0 & 84.5 & 95.5 & 97.6 & 45.1 & 46.7 \\
                                            & \texttt{7B}       & \textbf{79.5} & \textbf{85.3} & 96.9 & \textbf{98.3} & 47.7 & \textbf{48.8} \\
                                            & \texttt{20B}      & 76.8 & 83.6 & \textbf{97.1} & \textbf{98.3} & \textbf{48.1} & 48.6 \\
            \bottomrule
        \end{NiceTabular}
    }
    \caption{
        Sentence- and character-level $F_1$ scores of different model families and sizes. All LLMs are combined with ReLM. 
    }
    \label{tab:ablation:size}
\end{table}

\subsection{Case Study}

Figure~\ref{fig:examples} presents comparative cases demonstrating our approach's effectiveness in error correction.

In Figure~\ref{fig:examples:correct_both}, while ReLM provides partial corrections and the LLM fails to detect the error, our mixture approach achieves complete error resolution. This is because compared to \citet{zhou-etal-2024-llm-csc}, we reduce the weight of the distortion model in Eq.~\eqref{eq:final_score}, which allows the language model to play a greater role.
Figure~\ref{fig:examples:correct_llm} illustrates a typical under-correction scenario. Here, the LLM's under-correction of ``驾驶者'' (driver) is successfully revised to ``驾驶证'' (driver's license) through ReLM's complementary intervention. 
The third case in Figure~\ref{fig:examples:correct_relm} reveals an over-correction scenario where ReLM introduces an erroneous modification, which is effectively identified and corrected by the LLM through beam search decoding.

In summary, LLMs demonstrate superior fluency preservation, while small models are better at making accurate corrections. The synergistic integration of both capabilities through our framework yields optimal correction results. 

\input{figures/alpha_beta.tex}
\section{Discussion}
Following \citet{zhou-etal-2024-llm-csc}, we evaluate our approach on three benchmark datasets: rSIGHAN15, ECSPell-\textit{Odw}, and LEMON-\textit{Nov}.

\subsection{Impact of Different LLM Families and Model Sizes}
\label{sec:ablation:family_size}

To systematically analyze how LLM families and model sizes affect correction performance, we conduct comprehensive experiments with multiple open-source LLMs scaled from 0.5B to 20B parameters, including \texttt{Baichuan2} (7B-13B), \texttt{Qwen2.5} (0.5B-14B), and \texttt{InternLM2.5} (1.8B-20B). All models are integrated with ReLM and evaluated across all three datasets. Table~\ref{tab:ablation:size} reveals two key findings: 
\begin{asparaitem}[$\bullet$]
    \item Within the same LLM family, larger models do not necessarily lead to better performance. For example, on rSIGHAN15, the correction performance of all LLMs does not show an increasing trend with the growth of model size. 
    \item Different LLM families exhibit varying strengths across different domains. For example, \texttt{InternLM2.5} outperforms \texttt{Qwen2.5} on the rSIGHAN15 dataset, whereas the opposite is true for LEMON-\textit{Nov}. Overall, the combination of ReLM and \texttt{Baichuan2} shows more stable correction performance.
\end{asparaitem}

\begin{table}[tb!]
    \centering
    \scalebox{0.85}{
        \setlength{\tabcolsep}{2.6pt}%
        \renewcommand{\arraystretch}{0.92}
        \begin{NiceTabular}{lcc>{\columncolor{figure_light_blue!40}}c|cc>{\columncolor{figure_light_blue!40}}c|cc>{\columncolor{figure_light_red!6}}c}
            \toprule
            \rowcolor[gray]{1.0}
            \Block[l]{1-1}{\textbf{Model}}               & S-P\bgood  & S-R\bgood  & \textbf{S-F}\bgood & C-P\bgood  & C-R\bgood  & \textbf{C-F}\bgood & \textbf{FPR}\sgood \\
            \midrule
            \rowcolor[gray]{.95}
            \Block[l]{1-8}{\textbf{rSIGHAN15}}  &                    &            &            &                    &            &            &                    \\
            \texttt{ReLM}                                  & \wm76.8 	& \wm73.8 	& \wm75.2  	& \wm87.9 	& \wm78.0 	& \wm82.6  	& \wm\wz8.3             \\
            \texttt{BC2}                                   & \wm67.1 	& \wm55.8 	& \wm61.0   & \wm78.7 	& \wm61.0 	& \wm68.7     & \wm\wz8.3             \\
            \midrule
            \texttt{Ours}                                   & \wm\textbf{83.5} 	& \wm\textbf{76.1} 	& \wm\textbf{79.6}  	& \wm\textbf{93.8} 	& \wm78.3 	& \wm\textbf{85.4}  	& \wm\wz\textbf{4.1}          \\
            % \hdashedline
            \ \ \texttt{-\,DM}                            & \wz\ewm4.3         & \wz\ewm0.2   & \wz\ewm2.1 & \wz\ewm4.2         & \textbf{\wz\ewp1.3}   & \wz\ewm1.1 & \wz\ewp3.4         \\
            \ \ \texttt{-\,FR}                            & \wz\ewm3.6 & \wz\ewm0.9 & \wz\ewm2.1 & \wz\ewm3.1 & \wz\ewp0.3 & \wz\ewm1.2 & \wz\ewp2.3 \\
            \ \ \texttt{-\,both}                            & \wz\ewm4.7 & \wz\ewm0.5 & \wz\ewm2.5 & \wz\ewm4.6 & \wz\ewp1.1 & \wz\ewm1.4 & \wz\ewp3.8 \\
            % \midrule
            \midrule
            \rowcolor[gray]{.95}
            \Block[l]{1-8}{\textbf{ECSpell}-\textit{Odw}}   &                    &            &            &                    &            &            &                    \\
            \texttt{ReLM}                                  & \wm89.4          & \wm91.5  & \wm90.4  & \wm93.1          & \wm95.7  & \wm94.4    & \wm\wz5.8            \\
            \texttt{BC2}                                   & \wm92.0          & \wm89.1  & \wm90.5  & \wm95.0          & \wm92.6  & \wm93.8    & \wm\wz1.6            \\
            \midrule
            \texttt{Ours}                                  & \wm97.2          & \wm95.7  & \wm96.5  & \wm\textbf{98.7}          & \wm97.7  & \wm\textbf{98.2}    & \wm\wz\textbf{0.8}            \\
            % \hdashedline
            \ \ \texttt{-\,DM}                           & \wz\ewm5.2 	& \wz\ewm1.5 	& \wz\ewm3.4 	& \wz\ewm3.2 	& \wz\ewm0.8 	& \wz\ewm2.0 	& \wz\ewp3.3            \\
            \ \ \texttt{-\,FR}                            & \textbf{\wz\ewp0.4} 	& \textbf{\wz\ewp0.4} 	& \textbf{\wz\ewp0.4} 	& \wz\ewm0.2 	& \textbf{\wz\ewp0.3} 	& \textbf{\wz\ewp0.0} 	& \textbf{\wz\ewp0.0} \\
            \ \ \texttt{-\,both}                          & \wz\ewm4.4 	& \wz\ewm0.8 	& \wz\ewm2.7 	& \wz\ewm3.0 	& \wz\ewm0.5 	& \wz\ewm1.7 	& \wz\ewp3.3 \\
            % \midrule
            \midrule
            \rowcolor[gray]{.95}
            \Block[l]{1-8}{\textbf{Lemon}-\textit{Nov}} &                    &            &            &                    &            &            &                    \\
            \texttt{ReLM}                                  & \wm46.3            & \wm32.2    & \wm38.0    & \wm48.9            & \wm31.0    & \wm37.9    & \wm17.6            \\
            \texttt{BC2}                                   & \wm49.8            & \wm37.0    & \wm42.4    & \wm53.5            & \wm42.0    & \wm47.1    & \wm14.9            \\
            \midrule
            \texttt{Ours}                                  & \wm\textbf{64.0}            & \wm41.0    & \wm\textbf{50.0}    & \wm\textbf{67.2}            & \wm40.3    & \wm50.4    & \wm\wz\textbf{9.8}          \\
            % \hdashedline
            \ \ \texttt{-\,DM}    & \ewm15.8    & \wz\ewm4.2    & \wz\ewm8.2    & \ewm16.1    & \wz\ewm4.3    & \wz\ewm8.1    & \wz\ewp7.8 \\
            \ \ \texttt{-\,FR}    & \wz\ewm6.2  & \textbf{\wz\ewp3.1}    & \textbf{\wz\ewp0.0}    & \wz\ewm6.2  & \textbf{\wz\ewp4.5}    & \textbf{\wz\ewp1.3}    & \wz\ewp4.3 \\
            \ \ \texttt{-\,both}  & \ewm16.4    & \wz\ewm3.0    & \wz\ewm7.8    & \ewm16.5    & \wz\ewm2.6    & \wz\ewm7.1    & \wz\ewp8.6 \\
            \bottomrule
        \end{NiceTabular}
    }
    \caption{
        Ablation results of distortion model (DM) and faithfulness reward (FR) on ReLM + \texttt{Baichuan2(7B)}.
        ``\texttt{ReLM}'' and ``\texttt{BC2}'' represent using ReLM and \texttt{Baichuan2(7B)} model alone respectively.
        ``\texttt{-both}'' represents that we remove the intervention of both DM and FR on the results.
    }
    \label{tab:ablation:DM}
\end{table}

\subsection{Impact of Hyperparameters}
We conduct an analysis of two critical hyperparameters in Eq.~\eqref{eq:final_score}: the distortion model weight $\alpha$ and the small BERT-based model weight $\beta$.
Figure~\ref{fig:ablation:alpha_beta} demonstrates their effects on three 7B-scale LLM families evaluated on rSIGHAN15.
For $\alpha$, most LLMs achieve optimal performance around 0.5. Compared with \texttt{Baichuan2}, both \texttt{Qwen2.5} and \texttt{InternLM2.5} require higher $\beta$ values for peak performance. This difference likely stems from their weaker performance on rSIGHAN15 compared to \texttt{Baichuan2}. This hypothesis is further supported by experiments on different test sets. 
On ECSpell-\textit{Odw}, where the small and large models demonstrate closer correction capabilities, we observe greater flexibility in $\beta$ selection. Conversely, for LEMON, the superior performance of LLMs reduces the significance of small models.

It is important to note that in Table~\ref{tab:main_results}, the weights are fixed ($\alpha=0.5$, $\beta=0.9$). In practice, tuning these weights can further enhance model performance. Importantly, our approach consistently surpasses the ReLM baseline across all hyperparameter settings, highlighting its robustness.

\subsection{Impact of Distortion Model and Faithfulness Reward}

Table~\ref{tab:ablation:DM} investigates the individual contributions of the distortion model (DM) and faithfulness reward (FR) through an ablation study. The results show that while the removal of DM leads to performance degradation, our approach still outperforms the ReLM baseline. This demonstrates that even when stripped of LLM-specific correction strategies, an LLM retaining core language modeling capability can still provide effective corrective guidance to the small model.

Moreover, removing the DM and FR has a larger impact on Precision than on Recall. This is because the DM’s similarity information and FR's constraints help the model avoid unnecessary modifications.

\subsection{Impact of Small BERT-based Models}
\begin{table}[tb!]
    \centering
    \scalebox{0.85}{
        \setlength{\tabcolsep}{2.6pt}%
        \renewcommand{\arraystretch}{0.92}
        \begin{NiceTabular}{lcc>{\columncolor{figure_light_blue!40}}c|cc>{\columncolor{figure_light_blue!40}}c|cc>{\columncolor{figure_light_red!6}}c}
            \toprule
            \rowcolor[gray]{1.0}
            \Block[l]{2-1}{\textbf{Model}}  & \Block[c]{1-7}{\textbf{rSIGHAN15}}     & & & & & & \\                 
            \rowcolor[gray]{1.0}  & S-P\bgood  & S-R\bgood  & \textbf{S-F}\bgood & C-P\bgood  & C-R\bgood  & \textbf{C-F}\bgood & \textbf{FPR}\sgood \\
            \midrule
            \rowcolor[gray]{.95}
            \Block[c]{1-8}{\texttt{LLMs} \cite{zhou-etal-2024-llm-csc}}   &                    &            &            &                    &            &            &                    \\
            \texttt{BC2}                          & \wm67.1 	& \wm55.8 	& \wm61.0   & \wm78.7 	& \wm61.0 	& \wm68.7     & \wm\wz8.3            \\
            \midrule
            \rowcolor[gray]{.95}
            \Block[c]{1-8}{\texttt{Previous\;SOTAs \& Ours}} &                    &            &            &                    &            &            &                    \\
            \texttt{ReaLiSe}                               & \wm75.7 	& \wm70.2 	& \wm72.9 	& \wm83.4 	& \wm73.9 	& \wm78.4 	& \wm\wz8.1             \\
            \texttt{\quad +\;BC2}                          & \wm80.5 	& \wm72.0 	& \wm76.0 	& \wm92.9 	& \wm74.6 	& \wm82.7 	& \wm\wz5.2             \\
            \hdashedline
            \texttt{BERT}                                  & \wm75.3 	& \wm74.9 	& \wm75.1 	& \wm86.2 	& \wm\textbf{79.6} 	& \wm82.8 	& \wm10.8             \\
            \texttt{\quad +\;BC2}                          & \wm82.6 	& \wm74.4 	& \wm78.3 	& \wm93.3 	& \wm78.4 	& \wm\textbf{85.2} 	& \wm\wz3.5             \\
            \hdashedline
            \texttt{ReLM}                                  & \wm76.8 	& \wm73.8 	& \wm75.2 	& \wm87.9 	& \wm78.0 	& \wm82.6 	& \wm\wz8.3             \\
            \texttt{\quad +\;BC2}                          & \wm83.1 	& \wm\textbf{75.6} 	& \wm79.2 	& \wm\textbf{93.8} 	& \wm78.1 	& \wm\textbf{85.2} 	& \wm\wz4.1             \\
            \hdashedline
            \texttt{SCOPE}                                 & \wm78.7 	& \wm73.5 	& \wm76.0 	& \wm83.9 	& \wm76.7 	& \wm80.1 	& \wm\wz7.0             \\
            \texttt{\quad +\;BC2}                          & \wm\textbf{84.9} 	& \wm75.4 	& \wm\textbf{79.9} 	& \wm\textbf{93.8} 	& \wm77.2 	& \wm84.7 	& \wm\wz\textbf{3.3}             \\
            \bottomrule
        \end{NiceTabular}
    }
    \caption{
        Ablation results of SOTA BERT-based models combined with \texttt{Baichuan2(7B)}.
    }
    \label{tab:ablation:BM}
\end{table}

To evaluate the impact of small models on experimental outcomes, we conducted experiments with several classic and high-performing BERT-based models in combination with \texttt{Baichuan2(7B)}. As demonstrated in Table~\ref{tab:ablation:BM} and Table~\ref{tab:sighan15}, the integration with LLMs leads to substantial performance improvements across all small models. Notably, this enhancement is reflected in both Precision and Recall metrics, while simultaneously reducing the FPR. Furthermore, our analysis reveals a strong positive correlation between the performance of the selected small models and the overall correction performance of our mixture approach.

\subsection{Conflicts Between Our Method and In-context Learning (ICL) Technique}
\begin{table}[tb!]
    \centering
    \scalebox{0.85}{
        \setlength{\tabcolsep}{2.6pt}%
        \renewcommand{\arraystretch}{0.92}
        \begin{NiceTabular}{lcc>{\columncolor{figure_light_blue!40}}c|cc>{\columncolor{figure_light_blue!40}}c|cc>{\columncolor{figure_light_red!6}}c}
            \toprule
            \rowcolor[gray]{1.0}
            \Block[l]{1-1}{\textbf{Model}}               & S-P\bgood  & S-R\bgood  & \textbf{S-F}\bgood & C-P\bgood  & C-R\bgood  & \textbf{C-F}\bgood & \textbf{FPR}\sgood \\
            \midrule
            \rowcolor[gray]{.95}
            \Block[l]{1-8}{\textbf{rSIGHAN15}}  &                    &            &            &                    &            &            &                    \\
            \texttt{BC2}                                  & \wm67.1 	& \wm55.8 	& \wm61.0   & \wm78.7 	& \wm61.0 	& \wm68.7     & \wm\wz8.3             \\
            \texttt{BC2}\rlap{*}                                   & \wm72.3 	& \wm45.1 	& \wm55.6   & \wm88.7 	& \wm45.2 	& \wm59.9     & \wm\wz2.7             \\
            
            \midrule
            \texttt{ReLM}                                  &           &   &   &           &   &     &             \\
            \ \ \texttt{+\,BC2}                           & \wm83.5 	& \wm\textbf{76.1} 	& \wm\textbf{79.6}  	& \wm93.8 	& \wm\textbf{78.3} 	& \wm\textbf{85.4}  	& \wm\wz4.1            \\
            \ \ \texttt{+\,BC2}\rlap{*}                            & \wz\textbf{83.9} 	& \wz65.0 	& \wz73.3 	& \wz\textbf{97.4} 	& \wz66.6 	& \wz79.1 	& \wz\wz\textbf{1.2}            \\
            \midrule
            \rowcolor[gray]{.95}
            \Block[l]{1-8}{\textbf{ECSpell}-\textit{Odw}}   &                    &            &            &                    &            &            &                    \\
            \texttt{BC2}                                   & \wm92.0          & \wm89.1  & \wm90.5  & \wm95.0          & \wm92.6  & \wm93.8    & \wm\wz1.6            \\
            \texttt{BC2}\rlap{*}                                  & \wm85.4          & \wm59.1  & \wm69.9  & \wm96.7          & \wm60.5  & \wm74.4    & \wm\wz1.2            \\
            \midrule
            \texttt{ReLM}                                  &           &   &   &           &   &     &            \\
            \ \ \texttt{+\,BC2}                           & \wm\textbf{97.2}          & \wm\textbf{95.7}  & \wm\textbf{96.5}  & \wm98.7          & \wm\textbf{97.7}  & \wm\textbf{98.2}    & \wm\wz0.8            \\
            \ \ \texttt{+\,BC2}\rlap{*}                   & \wm96.3          & \wm91.8  & \wm94.0  & \wm\textbf{99.2}          & \wm93.6  & \wm96.3    & \wm\wz\textbf{0.4}            \\        
            \midrule
            \rowcolor[gray]{.95}
            \Block[l]{1-8}{\textbf{Lemon}-\textit{Nov}} &                    &            &            &                    &            &            &                    \\
            \texttt{BC2}                                   & \wm49.8            & \wm37.0    & \wm42.4    & \wm53.5            & \wm\textbf{42.0}    & \wm47.1    & \wm14.9            \\
            \texttt{BC2}\rlap{*}                                  & \wm61.0          & \wm19.0  & \wm29.0  & \wm64.5          & \wm19.6  & \wm30.0    & \wm\wz4.5           \\
            \midrule
            \texttt{ReLM}                                  &           &   &   &           &   &     &          \\
            \ \ \texttt{+\,BC2}    & \wm64.0            & \wm\textbf{41.0}    & \wm\textbf{50.0}    & \wm67.2            & \wm40.3    & \wm\textbf{50.4}    & \wm\wz9.8 \\
            \ \ \texttt{+\,BC2}\rlap{*}    & \wm\textbf{83.3}            & \wm23.3    & \wm36.4    & \wm\textbf{84.0}            & \wm21.8    & \wm34.6    & \wm\wz\textbf{1.5}            \\
            \bottomrule
        \end{NiceTabular}
    }
    \caption{
        Performance of \texttt{Baichuan2(7B)}. ``*'' means using the chat version of the LLM. 
    }
    \label{tab:ablation:ICL}
\end{table}

Our proposed method treats the LLM as a pure language model, utilizing its ability to generate correct and fluent text to assist the small model in error correction. When using ICL technique, the output of the LLM itself would become biased, and the settings for fluent text generation would be disrupted, leading to reduced performance. Therefore, the LLMs we use are the base version, meaning they have only undergone pre-training and have not been subject to Instruction Fine-Tuning (IFT), so they possess almost no conversational abilities. As a result, our approach conflicts with ICL technique. We attempted to replace the LLM with the chat version of \texttt{Baichuan2(7B)}, and the results are presented in Table~\ref{tab:ablation:ICL}.

The results show that ICL technique is not compatible with our approach. We found that the performance loss primarily affects Recall, while Precision sometimes even improves, which aligns with our previous analysis. ICL biases the LLM's output towards error correction, weakening its ability to model fluency, thus preserving Precision but significantly reducing Recall. The prompt used in our experiments is shown in Figure~\ref{fig:prompt_examples}.

\begin{table}[tb!]
    \centering
    \scalebox{0.85}{
        \setlength{\tabcolsep}{2.6pt}%
        \renewcommand{\arraystretch}{0.92}
        \begin{NiceTabular}{lccccc}
            \toprule
            \rowcolor[gray]{1.0}
            \Block[l]{1-1}{\textbf{Model}}               & SaP  & SiP  & SiS & Others  & Total \\
            \midrule
            \rowcolor[gray]{.95}
            \Block[l]{1-6}{\textbf{rSIGHAN15}}  &                    &            &            &                    &            \\
            \texttt{ReLM}                                  & \wm85.2 	& \wm84.6 	& \wm83.4   & \wm55.3 	& \wm82.6 	\\
            \texttt{BC2}                                 & \wm74.5 	& \wm67.7 	& \wm68.8   & \wm12.7 	& \wm68.7 	\\
            \texttt{ReLM\,+\,BC2}                           & \wm\textbf{87.2} 	& \wm\textbf{85.8} 	& \wm\textbf{83.5}  	& \wm\textbf{60.3} 	& \wm\textbf{85.4} 	\\
            \midrule
            \rowcolor[gray]{.95}
            \Block[l]{1-6}{\textbf{ECSpell}-\textit{Odw}}   &                    &            &            &                    &         \\
            \texttt{ReLM}                                  & \wm94.2 	& \wm86.3 	& \wm96.2   & \wm50.8 	& \wm90.6 	\\
            \texttt{BC2}                                  & \wm96.1 	& \wm88.1 	& \wm96.3   & \wm48.0 	& \wm93.8 	\\
            \texttt{ReLM\,+\,BC2}                           & \wm\textbf{98.8} 	& \wm\textbf{98.2} 	& \wm\textbf{99.2}  	& \wm\textbf{85.7} 	& \wm\textbf{98.3} 	\\        
            \midrule
            \rowcolor[gray]{.95}
            \Block[l]{1-6}{\textbf{Lemon}-\textit{Nov}} &                    &            &            &                    &            \\
            \texttt{ReLM}                                  & \wm49.7 	& \wm34.7 	& \wm43.8   & \wm13.6 	& \wm37.9 	\\
            \texttt{BC2}                                 & \wm60.5 	& \wm38.9 	& \wm54.0   & \wm\wz1.9 	& \wm47.1 	\\
            \texttt{ReLM\,+\,BC2}                           & \wm\textbf{61.2} 	& \wm\textbf{45.2} 	& \wm\textbf{54.2}  	& \wm\textbf{14.2} 	& \wm\textbf{50.4} 	\\
            \bottomrule
        \end{NiceTabular}
    }
    \caption{
        Character-level $F_1$ scores of different error types. ``SaP'', ``SiP'', and ``SiS'' denote the three types of spelling errors: same pinyin, similar pinyin, and similar shape, respectively.
    }
    \label{tab:error_types}
\end{table}

\subsection{Performance of Different Error Types}
We further broke down the character-level correction metrics with the error types categorization strategy from the distortion model. This allowed us to obtain specific performance results for each error type, such as same pinyin, similar pinyin, similar shape, and others, as shown in Table~\ref{tab:error_types}.

The results show that our mixture approach improves the F1 score across all error types, with simultaneous improvements in both Precision and Recall, which demonstrates the robustness of our method. It is important to note that the "Others" error type is the most challenging to correct in the CSC task. However, our method achieves significant improvements in this category, demonstrating that it effectively leverages the complementary strengths of LLMs and BERT-based models.

\subsection{Impact of Beam Size}
\begin{figure*}[tb!]
    \centering
    \captionsetup[subfigure]{skip=-1pt, margin=5pt}
    \begin{tikzpicture}[
            legend/.style={
                    fill=white,
                    font=\footnotesize,
                    inner sep=2pt,
                    minimum width=0.8cm,
                    text opacity=1.0,
                    fill opacity=1.0,
                },
            trim left
        ]
        \centering
        \begin{groupplot}[
                group style={
                        group size=3 by 1,
                        x descriptions at=edge bottom,
                        horizontal sep=0.7cm,
                        vertical sep=0.1cm,
                    },
                width=0.38\linewidth,
                height=0.28\linewidth,
                % width=1.0\linewidth,  % 从 0.38 减小
                % height=0.7\linewidth,  % 调整以保持宽高比
                xlabel={Beam Size},
                xmajorgrids=true,
                xtick={2, 4, 6, 8, 10, 12},
                xticklabels={2, 4, 6, 8, 10, 12},
                /tikz/font=\scriptsize,
                ylabel shift=-4pt,
                xlabel shift=-4pt,
                yticklabel shift=-2pt,
                xticklabel shift=-1pt,
                legend style={
                        at={(0.98,0.02)},
                        anchor=south east,
                        font=\tiny,
                        align=left,
                        nodes={anchor=west}
                    },
            ]
            \nextgroupplot[ymin=76,ymax=87.3, ytick={76, 78, 80, 82, 84, 86}]
            \addplot+ [mark=*, draw=figure_blue, thick,
                mark size=1.25pt,
                mark options={fill=figure_blue, fill opacity=1.0, solid},
                opacity=1.0,
            ] coordinates {
                (2, 77.3) (4, 77.8) (6, 77.9) (8, 78.4) (10, 78.6) (12, 79.2)
            };
            \addplot+ [mark=square*, draw=figure_red, thick,
                mark size=1.25pt,
                mark options={fill=white, fill opacity=1.0, solid},
                opacity=1.0,
            ] coordinates {
                (2, 77.1) (4, 77.6) (6, 77.2) (8, 77.5) (10, 78.1) (12, 77.9)
            };
            \addplot+ [mark=triangle*, draw=figure_green, thick,
                mark size=1.25pt,
                mark options={fill=white, fill opacity=1.0, solid},
                opacity=1.0,
            ] coordinates {
                (2, 76.6) (4, 77.8) (6, 78.4) (8, 79.6) (10, 79.5) (12, 79.5)
            };
            \addplot+ [mark=*, draw=figure_blue, thick, dashed,
                mark size=1.25pt,
                mark options={fill=figure_blue, fill opacity=1.0, solid},
                opacity=1.0,
            ] coordinates {
                (2, 83.6) (4, 84.1) (6, 84.2) (8, 84.4) (10, 84.6) (12, 85.2)
            };
            \addplot+ [mark=square*, draw=figure_red, thick, dashed,
                mark size=1.25pt,
                mark options={fill=white, fill opacity=1.0, solid},
                opacity=1.0,
            ] coordinates {
                (2, 83.3) (4, 84.4) (6, 84.0) (8, 84.3) (10, 84.4) (12, 84.2)
            };
            \addplot+ [mark=triangle*, draw=figure_green, thick, dashed,
                mark size=1.25pt,
                mark options={fill=white, fill opacity=1.0, solid},
                opacity=1.0,
            ] coordinates {
                (2, 83.6) (4, 84.7) (6, 84.8) (8, 85.4) (10, 85.4) (12, 85.3)
            };
            
            \nextgroupplot[ymin=94.5,ymax=99.3, ytick={95, 96, 97, 98, 99}]
            \addplot+ [mark=*, draw=figure_blue, thick,
                mark size=1.25pt,
                mark options={fill=figure_blue, fill opacity=1.0, solid},
                opacity=1.0,
            ] coordinates {
                (2, 95.1) (4, 95.9) (6, 96.5) (8, 96.9) (10, 96.9) (12, 96.5)
            };
            \addplot+ [mark=square*, draw=figure_red, thick,
                mark size=1.25pt,
                mark options={fill=white, fill opacity=1.0, solid},
                opacity=1.0,
            ] coordinates {
                (2, 95.7) (4, 95.7) (6, 95.9) (8, 95.9) (10, 95.7) (12, 95.9)
            };
            \addplot+ [mark=triangle*, draw=figure_green, thick,
                mark size=1.25pt,
                mark options={fill=white, fill opacity=1.0, solid},
                opacity=1.0,
            ] coordinates {
                (2, 94.9) (4, 96.7) (6, 96.9) (8, 97.3) (10, 96.9) (12, 96.9)
            };
            \addplot+ [mark=*, draw=figure_blue, thick, dashed,
                mark size=1.25pt,
                mark options={fill=figure_blue, fill opacity=1.0, solid},
                opacity=1.0,
            ] coordinates {
                (2, 97.4) (4, 97.9) (6, 98.2) (8, 98.3) (10, 98.3) (12, 98.2)
            };
            \addplot+ [mark=square*, draw=figure_red, thick, dashed,
                mark size=1.25pt,
                mark options={fill=white, fill opacity=1.0, solid},
                opacity=1.0,
            ] coordinates {
                (2, 97.7) (4, 97.7) (6, 97.7) (8, 97.7) (10, 97.6) (12, 97.7)
            };
            \addplot+ [mark=triangle*, draw=figure_green, thick, dashed,
                mark size=1.25pt,
                mark options={fill=white, fill opacity=1.0, solid},
                opacity=1.0,
            ] coordinates {
                (2, 97.4) (4, 98.2) (6, 98.3) (8, 98.5) (10, 98.3) (12, 98.3)
            };

            \nextgroupplot[ymin=42.5,ymax=52.3, ytick={43, 45, 47, 49, 51}]
            \addplot+ [mark=*, draw=figure_blue, thick,
                mark size=1.25pt,
                mark options={fill=figure_blue, fill opacity=1.0, solid},
                opacity=1.0,
            ] coordinates {
                (2, 45.3) (4, 47.7) (6, 49.0) (8, 49.8) (10, 50.0) (12, 49.8)
            };
            \addplot+ [mark=square*, draw=figure_red, thick,
                mark size=1.25pt,
                mark options={fill=white, fill opacity=1.0, solid},
                opacity=1.0,
            ] coordinates {
                (2, 44.6) (4, 47.3) (6, 48.6) (8, 48.8) (10, 49.3) (12, 49.4)
            };
            \addplot+ [mark=triangle*, draw=figure_green, thick,
                mark size=1.25pt,
                mark options={fill=white, fill opacity=1.0, solid},
                opacity=1.0,
            ] coordinates {
                (2, 43.6) (4, 46.3) (6, 46.7) (8, 46.7) (10, 47.2) (12, 47.7)
            };
            \addplot+ [mark=*, draw=figure_blue, thick, dashed,
                mark size=1.25pt,
                mark options={fill=figure_blue, fill opacity=1.0, solid},
                opacity=1.0,
            ] coordinates {
                (2, 45.7) (4, 48.0) (6, 49.0) (8, 50.0) (10, 50.4) (12, 50.3)
            };
            \addplot+ [mark=square*, draw=figure_red, thick, dashed,
                mark size=1.25pt,
                mark options={fill=white, fill opacity=1.0, solid},
                opacity=1.0,
            ] coordinates {
                (2, 45.3) (4, 48.2) (6, 49.5) (8, 50.1) (10, 50.5) (12, 50.6)
            };
            \addplot+ [mark=triangle*, draw=figure_green, thick, dashed,
                mark size=1.25pt,
                mark options={fill=white, fill opacity=1.0, solid},
                opacity=1.0,
            ] coordinates {
                (2, 44.5) (4, 47.2) (6, 48.0) (8, 48.1) (10, 48.5) (12, 48.8)
            };
            \legend{BC2, QW2.5, IL2.5};
        \end{groupplot}
        \node[anchor=north, legend] at (group c1r1.north) {\textbf{rSIGHAN15}};
        \node[anchor=north, legend] at (group c2r1.north) {\textbf{ECSpell}-\textit{Odw}};
        \node[anchor=north, legend] at (group c3r1.north) {\textbf{LEMON}-\textit{Nov}};
    \end{tikzpicture}
    \caption{
        $F_1$ scores of different LLMs (model size is 7B) with varying beam sizes. All models are combined with ReLM. Solid lines show sentence-level results, while dashed lines show character-level results.
    }
    \label{fig:beam:size}
\end{figure*}
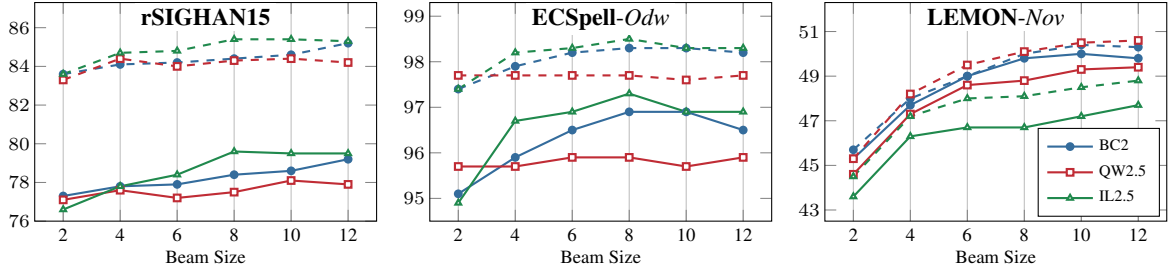

During beam search inference, beam size determines the number of candidate sequences maintained during decoding. While larger beam sizes enhance output diversity and better approximate the global optimal $\texttt{Score}(\boldsymbol{x}, \boldsymbol{y})$, this comes at the expense of decoding speed. 
Beam search can alleviate the limitation where generative models can only access previously generated tokens. In our approach, incorporating BERT-based models supplies additional contextual information to the LLMs, thereby reducing the dependence on beam search compared to using LLMs alone. 
As shown in Figure~\ref{fig:beam:size}, as beam size increases, the performance of all models generally shows an upward trend.

\subsection{Inference Speed}
\begin{table}[tb!]
    \setlength{\tabcolsep}{3pt}%
    \renewcommand{\arraystretch}{1.0}
    \centering
    \scalebox{1.0}{
        \begin{NiceTabular}{lcc}
            \toprule
            \rowcolor[gray]{1.0}
            \textbf{Model} & \textbf{Speed (ms/sent)} & \textbf{Slowdown} \\
            \midrule
            \texttt{BC2} & 1,276.0 & 1.00$\times$ \\
            \hdashedline
            \texttt{BERT} & 13.7 & -- \\
            \texttt{\quad +\;BC2} & 1,470.7 & 1.15$\times$ \\
            \hdashedline
            \texttt{ReLM} & 14.1 & -- \\
            \texttt{\quad +\;BC2} & 1,498.1 & 1.17$\times$ \\
            \bottomrule
        \end{NiceTabular}
    }
    \caption{The decoding time per sentence on rSIGHAN15. The LLM used is \texttt{Baichuan2(7B)}.}
    \label{tab:speed}
\end{table}

We compare the inference speeds of our approach and baseline models in Table~\ref{tab:speed}, with all experiments conducted on a single NVIDIA A100-PCIE-40GB GPU. The evaluation uses a batch size of 1 and a beam size of 12 for LLMs. The results reveal that our approach exhibits marginally slower inference speed compared to using LLMs alone, which can be attributed to the computational overhead introduced by integrating the small model's probability distribution at each beam search step.
    \section{Related Work}

\subsection{BERT-based Approaches}

Since the advent of BERT, BERT-based CSC models have shown strong correction capabilities. To better capture relationships between misspelled and correct characters, phonetic and glyph similarities have been incorporated into the encoder.
Techniques include using confusion sets for data augmentation \cite{Liu-etal-2021a-pretrain} and employing neural networks to encode phonetic and glyph features \cite{Xu-etal-2021-realise,huang-etal-2021-phmospell}. Additionally, new training objectives like pinyin prediction tasks have been developed to boost model performance \cite{Liu-etal-2021a-pretrain,Li-etal-2022-SCOPE,liang-etal-2023-disentangled}.

Other strategies involve modifying the model pipeline, for example, by adding a detection layer to enhance detection \cite{zhang-etal-2020-softmasked,Huang-etal-2023-module}; and using decoding intervention strategies to improve result selection \cite{wang-etal-2019-confusionset,Bao-etal-2020-confusionset,Lv-etal-2023-ECSpell,qiao-etal-2024-disc}.

\subsection{LLM Approaches}

In the era of LLMs, researchers have been actively exploring their potential in CSC tasks. A key consideration in these studies is whether LLMs require fine-tuning. \citet{li-2023-ineffectivenessLLMs} pioneered the exploration of prompt-based methods for correction tasks. They also experimented with SFT techniques, though the results were not satisfactory. \citet{dong-etal-2024-rich} enhanced the correction capabilities of LLMs by incorporating pinyin and radical information of Chinese characters into prompts. 
However, these prompt-based strategies struggle to maintain consistency between the prediction and input lengths.

Further advancements were made by \citet{li-etal-2024-cllm}, who replaced mixed tokenization with character-level tokenization. They fine-tune the LLMs to perform corrections in a character-by-character way, thereby resolving over 99\% of the length inconsistency issues.

In contrast to these methods, \citet{zhou-etal-2024-llm-csc} introduced a novel approach that is both prompt-free and training-free, treating LLMs as pure language models. This innovative approach opens a new research direction, aiming to fully leverage the intrinsic language modeling potential of LLMs.

\subsection{Mixture Approaches}

In the field of Grammatical Error Correction (GEC), model ensemble is commonly used. Multiple small models vote on edits to produce a final prediction \cite{zhang-etal-2022-mucgec}. \citet{zhou-etal-2023-improving-seq2seq} employed a language model, GPT-2, and a Grammatical Error Detection (GED) model, BART, as critics to dynamically guide the token selection in a Seq2Seq GEC model. They did not use an LLM due to the absence of a token-alignment method.

In the CSC domain, \citet{liu-etal-2024-arm} introduced an ARM method. Similar to the model ensemble technique, they replaced the voting process with an unfine-tuned small model that selects between the outputs of the LLM and a fine-tuned small model. Unlike their method, which focuses on post-processing predictions, our approach integrates both models during the beam search decoding process.
    \section{Conclusions}

In this paper, we propose a novel dynamic mixture approach that effectively combines small models and LLMs during the beam search decoding phase. By leveraging the strong error correction capabilities of fine-tuned BERT-based models and the language modeling strengths of LLMs, we achieve significant performance improvements. The advantage of not requiring fine-tuning of LLMs enhances the domain adaptability of our method. Experiments on mainstream public datasets demonstrate that our mixture approach achieves SOTA performance across multiple datasets.

\section*{Limitations}

\paragraph{Different Tasks.}
Our method is primarily designed for the CSC task. However, this mixture approach can also be adapted to other tasks, such as GEC, which is another type of text error correction, and even other areas that can be modeled simultaneously as classification and generation tasks.

\paragraph{Different Languages.}
Our method is currently limited to Chinese. However, for other languages, such as English, Korean, and Japanese, by modifying the token alignment between small models and LLMs, the same mixture approach can also achieve collaborative error correction.

\section*{Acknowledgements}
We sincerely thank the anonymous reviewers for their insightful comments and valuable efforts, which have significantly improved the quality and clarity of our work.

This work was supported by National Natural Science Foundation of China (Grant No. 62336006 and 62176173) and Project Funded by the Priority Academic Program Development of Jiangsu Higher Education Institutions.

    \bibliography{custom}
    % \bibliography{anthology,custom}
    \bibliographystyle{acl_natbib}

    % \appendix
    % \clearpage
    \appendix
\section{Details of Datasets}
\label{sec:appendix-datasets}
\paragraph{SIGHANs.}
The SIGHAN collection comprises three Chinese learner corpora (SIGHAN 13/14/15) \cite{wu-etal-2013-chinese,yu-etal-2014-overview,tseng-etal-2015-introduction}, which serve as standard benchmarks for CSC research. To address the noise and annotation errors in the original datasets, we adopt the revised version \cite{Yang-etal-2023-rSIGHAN} (\textbf{rSIGHANs}), which was manually re-annotated. Following prior setups, we combine SIGHAN datasets with Wang271K \cite{wang-etal-2018-hybrid} (containing 271K synthetic training instances) to form the composite training set.

\paragraph{LEMON.}
Containing over 22K sentences across seven distinct domains (game, encyclopedia, contract, medical care, car, novel, and news) \cite{Wu-etal-2023c-LEMON}, LEMON provides a cross-domain evaluation framework for CSC systems. Due to the absence of in-domain training sets, LEMON is employed to assess the cross-domain generalization capability of CSC models.

\paragraph{ECSpell.}
ECSpell consists of three small-scale datasets from the domains of law, medical treatment, and official document writing \cite{Lv-etal-2023-ECSpell}, offering domain-specific evaluation scenarios for CSC models.

\paragraph{CSCD-NS.}
CSCD-NS \cite{hu-etal-2024-cscd} contains 40K annotated samples sourced from real posts on Sina Weibo, effectively reflecting the real-world error correction performance of CSC models.

\paragraph{MCSCSet.}
Developed for medical domain applications \cite{jiang-etal-2022-mcscset}, this large-scale corpus contains approximately 200K professionally annotated sentences. The significant domain gap between medical texts and open-domain datasets makes MCSCSet particularly valuable for evaluating domain adaptation capabilities in CSC systems.

Detailed statistics are presented in Table~\ref{tab:detailed_statistics}.

\section{Details of Baselines}
\label{sec:appendix-baselines}
In our experiments, we selected four BERT-based models as baselines: BERT, ReaLiSe, SCOPE, and ReLM.
\begin{asparaitem}[$\bullet$]
    \item \textbf{BERT}: The vanilla BERT architecture implemented using the \texttt{bert-chinese-base}\footnote{\url{https://huggingface.co/bert-base-chinese}} pre-trained weights.
    \item \textbf{ReaLiSe}: This model employs GCN and CNN respectively to encode the pinyin and visual features (font images) of each Chinese character. These additional representations help the model capture the intrinsic phonetic and glyph relationships between characters. It uses \texttt{chinese-roberta-wwm-ext}\footnote{\url{https://huggingface.co/hfl/chinese-roberta-wwm-ext}} as its backbone.
    \item \textbf{SCOPE}: As one of the current SOTA CSC models, SCOPE incorporates an extra training task for character pronunciation prediction (CPP). The model is initialized from \texttt{ChineseBERT-base}\footnote{\url{https://huggingface.co/ShannonAI/ChineseBERT-base}}.
    \item \textbf{ReLM}: ReLM reframes the CSC task as a rephrasing problem rather than a tagging task. In practice, it still adopts non-autoregressive inference and remains essentially a BERT-based model, pre-trained and fine-tuned based on \texttt{bert-chinese-base}.
\end{asparaitem}

\begin{table}[tp!]
    \setlength{\tabcolsep}{12pt}
    \centering
    \scalebox{0.70}{
        \begin{tabular}{ll}
            \toprule
            \textbf{Source}                                                       \\
            \midrule
            \rowcolor[gray]{.9}
            From Test set: \\
            行政机关实施行政管理都应当公开，这是程序正\textcolor{figure_red}{档}原则的要求 \\
            \midrule
            \rowcolor[gray]{.9}
            From Training set: \\
            行政机关实施行政管理都应当公开，这是程序正当原则的要求 \\
            行政机关实施行政管理都应当公开，这是程序\textcolor{figure_red}{争}当原则的要求 \\
            行政机关实施行政管理都应当公开，这是程序正当原则的要求 \\

            行政机关实施行政管理都应当公开，这是程序\textcolor{figure_red}{止}当原则的要求 \\
            行政机关实\textcolor{figure_red}{私}行政管理都应当公开，这是程序正当原则的要求 \\
            \bottomrule
        \end{tabular}

    }
    \caption{
        Examples of ECSpell-\textit{Law}.
    }
    \label{tab:ecspell_example}
\end{table}
\begin{table}[tb!]
    \centering
    \scalebox{0.85}{
        \setlength{\tabcolsep}{2.6pt}%
        \renewcommand{\arraystretch}{0.92}
        \begin{NiceTabular}{lcc>{\columncolor{figure_light_blue!40}}c|cc>{\columncolor{figure_light_blue!40}}c|cc>{\columncolor{figure_light_red!6}}c}
            \toprule
            \rowcolor[gray]{1.0}
            \Block[l]{2-1}{\textbf{Model}}  & \Block[c]{1-7}{\textbf{SIGHAN15}}     & & & & & & \\                 
            \rowcolor[gray]{1.0}  & S-P\bgood  & S-R\bgood  & \textbf{S-F}\bgood & C-P\bgood  & C-R\bgood  & \textbf{C-F}\bgood & \textbf{FPR}\sgood \\
            \midrule
            \rowcolor[gray]{.95}
            \Block[c]{1-8}{\texttt{LLMs} \cite{zhou-etal-2024-llm-csc}}   &                    &            &            &                    &            &            &                    \\
            \texttt{BC2}                          & \wm61.2 	& \wm58.2 	& \wm59.7   & \wm69.0 	& \wm65.4 	& \wm67.2     & \wm14.3            \\
            \midrule
            \rowcolor[gray]{.95}
            \Block[c]{1-8}{\texttt{Previous\;SOTAs \& Ours}} &                    &            &            &                    &            &            &                    \\
            \texttt{ReaLiSe}                                & \wm75.9 	& \wm79.9 	& \wm77.8 	& \wm83.4 	& \wm83.8 	& \wm83.6 	& \wm12.0              \\
            \texttt{\quad +\;BC2}                          & \wm80.7 	& \wm82.4 	& \wm81.5 	& \wm87.2 	& \wm85.3 	& \wm86.3 	& \wm\wz9.5              \\
            \hdashedline
            \texttt{BERT}                                  & \wm75.2 	& \wm83.4 	& \wm79.1 	& \wm81.9 	& \wm89.2 	& \wm85.4 	& \wm14.1              \\
            \texttt{\quad +\;BC2}                          & \wm81.7 	& \wm84.1 	& \wm82.9 	& \wm86.8 	& \wm\textbf{89.6} 	& \wm\textbf{88.2} 	& \wm\wz7.9              \\
            \hdashedline
            \texttt{ReLM}                                  & \wm76.8 	& \wm83.9 	& \wm80.2 	& \wm83.2 	& \wm88.6 	& \wm85.8 	& \wm12.7              \\
            \texttt{\quad +\;BC2}                          & \wm82.8 	& \wm\textbf{85.9} 	& \wm84.3 	& \wm87.6 	& \wm88.7 	& \wm88.1 	& \wm\wz\textbf{7.8}              \\
            \hdashedline

            \texttt{SCOPE}                                 & \wm78.6 	& \wm83.5 	& \wm81.0 	& \wm83.3 	& \wm86.6 	& \wm84.9 	& \wm11.3              \\
            \texttt{\quad +\;BC2}                          & \wm\textbf{83.8} 	& \wm85.0 	& \wm\textbf{84.4} 	& \wm\textbf{88.1} 	& \wm87.6 	& \wm87.9 	& \wm\wz7.9              \\
            \texttt{\quad +\;ARM}                          & \wm79.5 	& \wm83.1 	& \wm81.3 	& \wm-- 	& \wm-- 	& \wm-- 	& \wm\wz--              \\
            \bottomrule
        \end{NiceTabular}
    }
    \caption{
        Sentence- and character-level results on the SIGHAN15 test set. 
        \texttt{ARM} represents the latest approach that integrates the results of small models and \texttt{GPT-3.5-Turbo} \cite{liu-etal-2024-arm}.
    }
    \label{tab:sighan15}
\end{table}

\begin{table*}[tb!]
    \centering
    % \subfloat[]{
    \setlength{\tabcolsep}{4.0pt}
    % \begin{NiceTabular}{lccc|c|c|ccc|c}
    \begin{NiceTabular}{p{8.7em}c|c|c|c|ccc}
        \toprule
        \rowcolor[gray]{.9}
        \textbf{Training Sets}    & \Block[c]{1-1}{\textbf{SIGHANs}} & \Block[c]{1-1}{\textbf{Wang271K}} & \Block[c]{1-1}{\textbf{CSCD-NS}} & \Block[c]{1-1}{\textbf{MCSCSet}} & \Block[c]{1-3}{\textbf{ECSpell}} &              &              \\
        \textbf{Subsets}                                         & -- & -- & -- & -- & \textit{Law}                     & \textit{Med} & \textit{Odw} \\
        \midrule
        \textbf{\#Sent.}                                      & 6,479                            & 271,329        & 30,000        & 157,194                         & 1,960 & 3,000 & 1,720                      \\
        \textbf{Avg. Length}                                  & 42.1                             & 42.6        & 57.4        & 10.9                        & 30.7                            & 50.2        & 41.2        \\
        \textbf{Avg. Error/Sent.}                          & \wz1.0                             & \wz1.4        & \wz0.5        & \wz0.9                        & \wz0.9                            & \wz0.8        & \wz0.9        \\
        \bottomrule
    \end{NiceTabular}
    % }\\
    \\[2pt]
    % \subfloat[]{
    \setlength{\tabcolsep}{4.0pt}
    % \begin{NiceTabular}{p{12.5em}ccccccc|c}
    \begin{NiceTabular}{lccc|c|c|ccc|c}
        \toprule
        \rowcolor[gray]{.9}
        \textbf{Test Sets}    & \Block[c]{1-3}{\textbf{rSIGHANs}} & & & \Block[c]{1-1}{\textbf{CSCD-NS}} & \Block[c]{1-1}{\textbf{MCSCSet}} & \Block[c]{1-3}{\textbf{ECSpell}} &              &             & \Block[c]{1-1}{\textbf{LEMON}} \\
        \textbf{Subsets}                                         & \textit{15} & \textit{14} & \textit{13} & -- & -- & \textit{Law}                     & \textit{Med} & \textit{Odw} & -- \\
        \midrule

        \textbf{\#Sent.}                                      & 1,100                            & 1,062        & 1,000        & 5,000                         & 19,650 & 500 & 500 & 500 & 22,252 \\
        \textbf{Avg. Length}                                  & 30.6                             & 50.0        & 74.3        & 57.6                        & 10.9                            & 29.7        & 49.6 & 40.5 & 35.4        \\

        \textbf{Avg. Error/Sent.}                          & \wz0.8                             & \wz0.9        & \wz1.5        & \wz0.5                        & \wz0.9                            & \wz0.8        & \wz0.7    & \wz0.8 & \wz0.5    \\
        \bottomrule
    \end{NiceTabular}
    % }
    \caption{
        Detailed statistics of all datasets used in our experiments.
    }
    \label{tab:detailed_statistics}
\end{table*}

\section{More Experiments}
\label{sec:appendix-more-experiments}
In this section, we present detailed results across different domains and the results of some other CSC models, such as ARM \cite{liu-etal-2024-arm} and C-LLM \cite{li-etal-2024-cllm}.
\begin{table}[tb!]
    \centering
    \setlength{\tabcolsep}{6.5pt}
    \begin{NiceTabular}{lccc}
        \toprule
        \rowcolor[gray]{.9}
        \textbf{Datasets}                                      & \Block[c]{1-3}{\textbf{ECSpell (cleaned)}} &               &               \\
        \textbf{Subsets}                     & \textit{Law}                      & \textit{Med}  & \textit{Odw}  \\
        \midrule
        \rowcolor[gray]{.95}
        \Block[c]{1-4}{\texttt{Previous\;SOTA}} &                                  &               &               \\
        \Block[l]{1-1}{\texttt{ReLM}}                                                                       
        & 71.0 & 70.3 & 77.7 \\
        \midrule
        \rowcolor[gray]{.95}
        \Block[c]{1-4}{\texttt{LLMs} \cite{zhou-etal-2024-llm-csc}}                                             &                                  &               &               \\
        \Block[l]{1-1}{\texttt{Baichuan2}}
        & 83.7                     & \textbf{82.0} & \textbf{90.5} \\
        \Block[l]{1-1}{\texttt{Qwen2.5}}
        & 83.7 & 73.1 & 89.1 \\
        \Block[l]{1-1}{\texttt{IL2.5}}
        & \textbf{85.8} & 66.2 & 89.0 \\
        \midrule
        \rowcolor[gray]{.95}
        \Block[c]{1-4}{\texttt{Ours}}                                             &                                  &               &               \\
        \Block[l]{1-1}{\texttt{ReLM\;+\;BC2}}
        & 86.4 & \textbf{87.9} & 93.1 \\
        \Block[l]{1-1}{\texttt{ReLM\;+\;QW2.5}}
        & \textbf{89.0} & 86.7 & 93.5 \\
        \Block[l]{1-1}{\texttt{ReLM\;+\;IL2.5}}
        & 86.4 & 87.1 & \textbf{93.7} \\
        \bottomrule
    \end{NiceTabular}
    \caption{
        Experiments on cleaned ECSpell.
    }
    \label{tab:ecspell_cleaned}
\end{table}

\begin{figure}[tb]
    \centering%
    \newtcolorbox{promptbox}[1]{%
        left=0pt,
        right=0pt,
        top=0pt,
        bottom=0pt,
        boxsep=6pt,
        % colback=json_blue!3,
        colframe=black,
        title={#1},
    }
    \begin{minipage}[b]{\columnwidth}
        \begin{promptbox}{\textbf{System and User Prompts for LLMs}}
            \large
            \scriptsize
            \textbf{\texttt{System Prompt:}}\\
            你是一个优秀的中文拼写纠错模型，中文拼写纠错模型即更正用户输入句子中的拼写错误。
            
            \textbf{\texttt{User Prompt:}}\\
            给你一句可能包含错别字的文本，你需要输出正确的文本。请不要有除预测文本以外的任何输出。
            
            示例：
            
            输入：消费者对事物的安全性问题越来越重视，消费有机食材在经济上已不再是顾虑
            
            输出：消费者对食物的安全性问题越来越重视，消费有机食材在经济上已不再是顾虑
            
            输入：它们曾今是朋友，但是现在却是对手
            
            输出：他们曾经是朋友，但是现在却是对手
            
            输入：近日办案民警终于分别在东莞、深圳两地抓获俩人
            
            输出：近日办案民警终于分别在东莞、深圳两地抓获俩人
            
            输入：[待纠错文本]
            
            输出：
        \end{promptbox}
    \end{minipage}
    \caption{Prompt template.}
    \label{fig:prompt_examples}
\end{figure}

\subsection{Experiments on Original SIGHAN15}
\label{sec:appendix-sighan15}

Following the experimental settings of previous studies, we use the original SIGHAN15 test set for comparison.\footnote{Since the dataset includes traditional characters, we use the preprocessed version by \citet{Xu-etal-2021-realise}.} In line with rSIGHAN15, we also adopt ReaLiSe and SCOPE, as shown in Table~\ref{tab:sighan15}. Evidently, our method significantly improves the performance of all baseline models, surpassing the current SOTA performance. 
We also show the latest approach that ensembles the results of small models and \texttt{GPT-3.5-Turbo} (SCOPE + ARM) \cite{liu-etal-2024-arm}, compared with which our method achieves a more effective performance improvement through deep integration of the small and large models during the inference stage.

\subsection{Detailed Results}
\label{sec:appendix-detailed-results}
\begin{table*}[tb!]
    \centering
    \setlength{\tabcolsep}{3.5pt}
    \renewcommand{\arraystretch}{1}
    \scalebox{0.85}{
        \begin{NiceTabular}{lccc|ccc|ccccccc}
            \toprule
            \rowcolor[gray]{.9}
            \textbf{Datasets}                   & \Block[c]{1-3}{\textbf{rSIGHANs}} &               &                                  & \Block[c]{1-3}{\textbf{ECSpell}} &               &               & \Block[c]{1-7}{\textbf{LEMON}} &               &               &               &               &               &               \\
            \textbf{Subsets}        & \textit{15} & \textit{14} & \textit{13}             & \textit{Law}                      & \textit{Med}  & \textit{Odw}  & \textit{Car}                   & \textit{Cot}  & \textit{Enc}  & \textit{Gam}  & \textit{Med}  & \textit{New}  & \textit{Nov}  \\
            \midrule
            \rowcolor[gray]{.95}
            \Block[c]{1-14}{\texttt{Previous\;SOTAs}} & & & &                                  &               &               &                                &               &               &               &               &               &               \\
            \Block[l]{1-1}{\texttt{BERT}}                                                                            
            & 75.1 	& 65.6 	& 71.6 & 95.9 & 88.0 & 89.1 & 52.0 & 63.8 & 45.3 & 32.9 & 50.8 & 56.0 & 35.8 \\
            \Block[l]{1-1}{\texttt{ReLM}}                                                                       
            & 75.2 	& 65.9 	& 74.4 & 96.2 & 90.2 & 90.4 & 53.6 & 67.7 & 47.7 & 34.6 & 53.9 & 58.8 & 38.0 \\
            \Block[l]{1-1}{\texttt{MDCSpell}$^\star$}                                         
            & -- 	& -- 	& -- & -- & -- & -- & 34.1 & 49.2 & 32.8 & 14.8 & 29.5 & 34.4 & 14.3 \\
            % \Block[l]{1-1}{\texttt{ReLM}$^\star$}
            % & -- 	& -- 	& -- & -- & -- & -- & 42.8 & 54.8 & 43.5 & 26.4 & 49.9 & 28.9 & 29.8 \\
            \midrule
            \rowcolor[gray]{.95}
            \Block[c]{1-14}{\texttt{LLMs} \cite{zhou-etal-2024-llm-csc}}  & & &                                            &                                  &               &               &                                &               &               &               &               &               &               \\
            \Block[l]{1-1}{\texttt{Baichuan2}}
            & \textbf{61.0} 	& \textbf{53.1} 	& \textbf{63.1} & 83.7                     & \textbf{82.0} & \textbf{90.5} & \textbf{54.2}                  & \textbf{63.3} & \textbf{51.4} & \textbf{36.9} & \textbf{60.6} & \textbf{63.9} & \textbf{42.4} \\
            \Block[l]{1-1}{\texttt{Qwen2.5}}
            & 58.3 	& 50.7 	& 57.9 & 83.7 & 73.1 & 89.1 & 48.2 & 59.9 & 46.5 & 34.7 & 54.2 & 59.5 & 37.1 \\
            \Block[l]{1-1}{\texttt{IL2.5}}
            & 55.6 	& 45.9 	& 56.0 & \textbf{85.8} & 66.2 & 89.0 & 44.4 & 54.7 & 45.2 & 33.0 & 50.2 & 57.2 & 32.1 \\
            \midrule
            \rowcolor[gray]{.95}

            \Block[c]{1-14}{\texttt{Mixture}}                 & & &                            &                                  &               &               &                                &               &               &               &               &               &               \\
            \Block[l]{1-1}{\texttt{MDCSpell\;+\;ARM}$^\star$} & -- 	& -- 	& -- & -- & -- & -- & 37.1 & 52.7 & 35.2 & 15.3 & 33.0 & 36.4 & 15.6 \\
            % \Block[l]{1-1}{\texttt{ReLM$^\star$\;+\;BC2}} & -- 	& -- 	& -- & -- & -- & -- & -- & -- & -- & 41.4 & -- & -- & -- \\
            \hdashedline
            \Block[l]{1-1}{\texttt{BERT\;+\;BC2}} & 78.3 	& 67.1 	& 72.1 & 95.7 & 93.8 & 93.6 & 56.0 & 68.0 & 49.4 & 41.2 & 63.9 & 68.7 & 47.6 \\
            \Block[l]{1-1}{\texttt{BERT\;+\;QW2.5}} & 77.7 	& \textbf{69.5} 	& 73.7 & 96.5 & 94.1 & 91.5 & 57.7 & 67.0 & 52.4 & 37.9 & 65.2 & 67.5 & 44.7 \\
            \Block[l]{1-1}{\texttt{BERT\;+\;IL2.5}} & 78.4 	& 67.7 	& 72.3 & 96.3 & 94.7 & 92.0 & 55.8 & 67.4 & 51.6 & 38.3 & 55.6 & 64.8 & 43.1 \\
            \hdashedline
            \Block[l]{1-1}{\texttt{ReLM\;+\;BC2}}
            & \textbf{79.6} 	& 67.6 	& 74.6 & \textbf{98.3} & \textbf{96.6} & 96.5 & 62.7 & \textbf{74.5} & 56.5 & \textbf{45.1} & 66.3 & \textbf{72.0} & \textbf{50.0} \\
            \Block[l]{1-1}{\texttt{ReLM\;+\;QW2.5}}
            & 77.9 	& 67.7 	& \textbf{76.0} & 98.1 & 95.9 & 95.9 & \textbf{63.6} & 71.2 & \textbf{57.4} & 44.0 & \textbf{66.7} & 71.4 & 49.4 \\
            \Block[l]{1-1}{\texttt{ReLM\;+\;IL2.5}}
            & 79.5 	& 68.3 	& 75.3 & 97.3 & 95.9 & \textbf{96.9} & 61.3 & 72.0 & 56.1 & 39.5 & 64.7 & 71.0 & 47.7 \\
            \bottomrule
        \end{NiceTabular}
    }
    \caption{
        Sentence-level $F_1$ (S-F) scores across different domains. ``$\star$'' indicates that the results are extracted from \citet{liu-etal-2024-arm} and the small model is trained on Wang271K + SIGHANs.
    }
    \label{tab:subsets_sf}
\end{table*}

\begin{table*}[tb!]
    \centering
    \setlength{\tabcolsep}{3.5pt}
    \renewcommand{\arraystretch}{1}
    \scalebox{0.85}{
        \begin{NiceTabular}{lccc|ccc|ccccccc}
            \toprule
            \rowcolor[gray]{.9}
            \textbf{Datasets}                   & \Block[c]{1-3}{\textbf{rSIGHANs}} &               &                                  & \Block[c]{1-3}{\textbf{ECSpell}} &               &               & \Block[c]{1-7}{\textbf{LEMON}} &               &               &               &               &               &               \\
            \textbf{Subsets}        & \textit{15} & \textit{14} & \textit{13}             & \textit{Law}                      & \textit{Med}  & \textit{Odw}  & \textit{Car}                   & \textit{Cot}  & \textit{Enc}  & \textit{Gam}  & \textit{Med}  & \textit{New}  & \textit{Nov}  \\
            \midrule
            \rowcolor[gray]{.95}
            \Block[c]{1-14}{\texttt{Previous\;SOTAs}} & & & &                                  &               &               &                                &               &               &               &               &               &               \\
            \Block[l]{1-1}{\texttt{BERT}}                                                                            
            & 82.8 	& 77.3 	& 85.7 & 97.4 & 93.3 & 93.2 & 52.7 & 65.3 & 46.1 & 35.6 & 52.0 & 57.4 & 36.3 \\
            \Block[l]{1-1}{\texttt{ReLM}}                                                                       

            & 82.6 	& 76.7 	& 86.1 & 93.6 & 94.4 & 94.4 & 54.3 & 67.4 & 48.1 & 37.9 & 54.9 & 60.5 & 37.9 \\
            \midrule
            \rowcolor[gray]{.95}

            \Block[c]{1-14}{\texttt{LLMs} \cite{zhou-etal-2024-llm-csc,li-etal-2024-cllm}}  & & &                                            &                                  &               &               &                                &               &               &               &               &               &               \\
            \Block[l]{1-1}{\texttt{Baichuan2}}
            & \textbf{68.7} 	& \textbf{64.9} 	& \textbf{76.8} & \textbf{88.0} & \textbf{93.8} & \textbf{93.8} & \textbf{58.8} & \textbf{65.3} & 56.3 & 40.9 & 61.6 & \textbf{66.8} & \textbf{47.1} \\
            \Block[l]{1-1}{\texttt{Qwen2.5}}
            & 66.1 	& 64.7 	& 75.0 & 81.6 & 93.0 & 93.0 & 54.1 & 62.5 & 52.4 & \textbf{41.8} & 56.6 & 63.2 & 43.8 \\
            \Block[l]{1-1}{\texttt{IL2.5}}
            & 64.6 	& 60.6 	& 73.4 & 78.3 & 92.4 & 92.4 & 49.9 & 58.1 & 51.6 & 37.5 & 53.5 & 61.0 & 38.6 \\
            \Block[l]{1-1}{\texttt{C-LLM}}
            & -- 	& -- 	& -- & -- & -- & -- & 57.5 & 60.4 & \textbf{56.5} & 38.0 & \textbf{65.3} & 64.5 & 43.9 \\
            \midrule
            \rowcolor[gray]{.95}

            \Block[c]{1-14}{\texttt{Ours}}                 & & &                            &                                  &               &               &                                &               &               &               &               &               &               \\
            \Block[l]{1-1}{\texttt{BERT\;+\;BC2}} 
            & 85.2 	& 78.3 	& 86.6 & 96.2 & 95.6 & 95.6 & 58.3 & 68.4 & 51.8 & 43.9 & 64.4 & 70.6 & 50.1 \\
            \Block[l]{1-1}{\texttt{BERT\;+\;QW2.5}} 
            & 84.6 	& \textbf{80.1} 	& 87.0 & 95.4 & 93.4 & 93.4 & 60.5 & 67.3 & 55.5 & 43.4 & 65.7 & 70.0 & 48.7 \\
            \Block[l]{1-1}{\texttt{BERT\;+\;IL2.5}} 
            & 84.8 	& 79.3 	& \textbf{87.4} & 92.7 & 94.4 & 94.4 & 57.4 & 68.0 & 52.3 & 41.5 & 56.0 & 66.2 & 44.0 \\
            \hdashedline

            \Block[l]{1-1}{\texttt{ReLM\;+\;BC2}}
            & \textbf{85.4} 	& 78.0 	& 85.7 & \textbf{97.8} & 98.2 & 98.2 & 62.5 & \textbf{74.2} & 56.7 & 47.8 & 66.5 & \textbf{73.1} & 50.4 \\
            \Block[l]{1-1}{\texttt{ReLM\;+\;QW2.5}}
            & 84.2 	& 77.1 	& 86.6 & 96.9 & 97.7 & 97.7 & \textbf{64.0} & 69.8 & \textbf{57.8} & \textbf{48.2} & \textbf{66.8} & 72.7 & \textbf{50.6} \\
            \Block[l]{1-1}{\texttt{ReLM\;+\;IL2.5}}
            & 85.3 	& 78.7 	& 86.8 & 97.5 & \textbf{98.3} & \textbf{98.3} & 62.2 & 71.8 & 57.4 & 44.1 & 65.1 & 72.1 & 48.8 \\
            \bottomrule
        \end{NiceTabular}
    }
    \caption{
        Character-level $F_1$ (C-F) scores across different domains.}
    \label{tab:subsets_cf}
\end{table*}

\begin{table*}[tb!]
    \centering
    \setlength{\tabcolsep}{3.5pt}
    \renewcommand{\arraystretch}{1}
    \scalebox{0.85}{
        \begin{NiceTabular}{lccc|ccc|ccccccc}
            \toprule
            \rowcolor[gray]{.9}
            \textbf{Datasets}                   & \Block[c]{1-3}{\textbf{rSIGHANs}} &               &                                  & \Block[c]{1-3}{\textbf{ECSpell}} &               &               & \Block[c]{1-7}{\textbf{LEMON}} &               &               &               &               &               &               \\
            \textbf{Subsets}        & \textit{15} & \textit{14} & \textit{13}             & \textit{Law}                      & \textit{Med}  & \textit{Odw}  & \textit{Car}                   & \textit{Cot}  & \textit{Enc}  & \textit{Gam}  & \textit{Med}  & \textit{New}  & \textit{Nov}  \\
            \midrule
            \rowcolor[gray]{.95}
            \Block[c]{1-14}{\texttt{Previous\;SOTAs}} & & & &                                  &               &               &                                &               &               &               &               &               &               \\
            \Block[l]{1-1}{\texttt{BERT}}                                                                            
            & 10.8 	& 13.0 	& 12.5 & \wz3.7 	& \wz5.1 	& \wz2.5 	& 12.2 	& \wz7.8 	& 13.8 	& 22.4 	& \wz8.5 	& \wz9.4 	& 17.3  \\
            \Block[l]{1-1}{\texttt{ReLM}}                                                                       
            & \wz8.3 	& 13.6 	& 12.6 & \wz6.5 	& \wz9.8 	& \wz5.8 	& 12.0 	& \wz4.9 	& 12.7 	& 20.6 	& \wz5.8 	& \wz8.4 	& 17.6  \\
            \midrule

            \rowcolor[gray]{.95}

            \Block[c]{1-14}{\texttt{LLMs} \cite{zhou-etal-2024-llm-csc}}  & & &                                            &                                  &               &               &                                &               &               &               &               &               &               \\
            \Block[l]{1-1}{\texttt{Baichuan2}}
            & \wz\textbf{8.3} 	& \textbf{15.7} 	& \textbf{22.3} & \wz4.9 	& \wz\textbf{8.7} 	& \wz\textbf{1.6} 	& \wz\textbf{6.9} 	& \wz\textbf{7.8} 	& \textbf{10.2} 	& \textbf{19.8} 	& \wz\textbf{3.4} 	& \wz\textbf{6.0} 	& \textbf{14.9}  \\
            \Block[l]{1-1}{\texttt{Qwen2.5}}
            & 10.2 	& 19.4 	& 24.2 & \wz\textbf{4.1} 	& 12.9 	& \wz2.9 	& 10.7 	& \wz6.7 	& 13.2 	& 23.9 	& \wz5.4 	& \wz9.2 	& 20.7  \\
            \Block[l]{1-1}{\texttt{IL2.5}}
            & 13.3 	& 21.4 	& 25.5 & \wz\textbf{4.1} 	& 14.0 	& \wz2.5 	& 13.0 	& 10.1 	& 15.2 	& 28.3 	& \wz7.9 	& \wz9.8 	& 24.8  \\
            \midrule
            \rowcolor[gray]{.95}

            \Block[c]{1-14}{\texttt{Ours}}                 & & &                            &                                  &               &               &                                &               &               &               &               &               &               \\
            \Block[l]{1-1}{\texttt{BERT\;+\;BC2}} 
            & \wz\textbf{3.5} 	& 10.0 	& \wz\textbf{8.7} & \wz\textbf{1.6} 	& \wz\textbf{2.6} 	& \wz\textbf{0.8} 	& \wz6.0 	& \wz2.3 	& \wz7.5 	& 14.5 	& \wz2.4 	& \wz4.2 	& 11.4  \\

            \Block[l]{1-1}{\texttt{BERT\;+\;QW2.5}} 
            & \wz3.7 	& \wz\textbf{8.7} 	& \wz\textbf{8.7} & \wz\textbf{1.6} 	& \wz4.4 	& \wz2.1 	& \wz8.4 	& \wz2.5 	& 10.9 	& 22.5 	& \wz3.1 	& \wz6.7 	& 16.9  \\
            
            \Block[l]{1-1}{\texttt{BERT\;+\;IL2.5}} 
            & \wz5.6 	& \wz9.3 	& \wz\textbf{8.7} & \wz2.0 	& \wz3.3 	& \wz1.3 	& \wz5.7 	& \wz2.2 	& \wz7.2 	& 14.1 	& \wz2.5 	& \wz3.5 	& 11.9  \\
            \hdashedline

            \Block[l]{1-1}{\texttt{ReLM\;+\;BC2}}
            & \wz4.1 	& 10.0 	& 13.4 & \wz2.4 	& \wz3.1 	& \wz\textbf{0.8} 	& \wz\textbf{4.5} 	& \wz\textbf{1.3} 	& \wz\textbf{6.6} 	& \textbf{11.6} 	& \wz2.3 	& \wz\textbf{3.1} 	& \wz\textbf{9.8}  \\

            \Block[l]{1-1}{\texttt{ReLM\;+\;QW2.5}}
            & \wz5.4 	& 11.0 	& 10.8 & \wz3.3 	& \wz4.5 	& \wz1.2 	& \wz6.4 	& \wz2.9 	& \wz9.2 	& 17.3 	& \wz\textbf{2.2} 	& \wz5.1 	& 12.2  \\
            
            \Block[l]{1-1}{\texttt{ReLM\;+\;IL2.5}}
            & \wz5.2 	& 10.6 	& 12.7 & \wz3.3 	& \wz3.8 	& \wz\textbf{0.8} 	& \wz6.8 	& \wz2.2 	& \wz9.1 	& 15.7 	& \wz3.3 	& \wz4.4 	& 13.9  \\
            \bottomrule
        \end{NiceTabular}
    }
    \caption{
        False positive rate (FPR) scores across different domains. }
    \label{tab:subsets_fpr}
\end{table*}

Due to space limitations, we present the detailed results of the subsets of rSIGHANs, ECSpell, and LEMON in Table~\ref{tab:subsets_sf}, Table~\ref{tab:subsets_cf}, and Table~\ref{tab:subsets_fpr}. 
We also provide the results of ARM and C-LLM in the corresponding tables. The results show that our method outperforms another mixture approach and the current SOTA SFT-based LLM strategy.

\subsection{Experiments on Cleaned ECSpell}
\label{sec:appendix-cleaned-ecspell}
Our analysis reveals that the strong performance of fine-tuned small models on ECSpell originates from substantial data leakage caused by homogeneous sentence pairs. Although ECSpell avoids including identical source-target sentence pairs, it introduces different synthesized errors for the same correct sentences that appear in both the training and test sets. These overlapping sentences account for 52.7\%, 19.3\%, and 28.2\% of the ECSpell-\textit{Law}/\textit{Med}/\textit{Odw} training sets, respectively. For example, the first sentence in the ECSpell-\textit{Law} test set appears five times in the training set as the same sentence, as illustrated in Table~\ref{tab:ecspell_example}.

Undoubtedly, the fine-tuning process of small models leads to results that are significantly higher than those of the LLMs without fine-tuning. Therefore, we removed the leaked sentences from the training sets and reorganized the experiments in Table~\ref{tab:ecspell_cleaned}. 
It is worth noting that all hyperparameter settings remain identical to those in the main experiments. Clearly, our approach continues to achieve significant and stable improvements.
In fact, due to the significantly reduced size of the training set, the performance of fine-tuned small models lags behind that of LLMs, and appropriately lowering the weight of the small model $\beta$ can yield even more significant performance improvements.

\end{CJK*}
\end{document}